\documentclass[lettersize,journal]{IEEEtran}
\usepackage{amsmath,amsfonts}
\usepackage{algorithmic}
\usepackage{algorithm}
\usepackage{array}
\usepackage[caption=false,font=normalsize,labelfont=sf,textfont=sf]{subfig}
\usepackage{textcomp}
\usepackage{stfloats}
\usepackage{url}
\usepackage{verbatim}
\usepackage{graphicx}
\usepackage[nocompress]{cite}
\usepackage{multirow}
\usepackage{bm}
\usepackage{rotating}
\usepackage{amsthm}
\usepackage[unicode]{hyperref}

\begin{document}
\bstctlcite{IEEEexample:BSTcontrol}

\title{Self-Gating Attention for Efficient Time Series Forecasting}

\author{Dezheng Wang, Tong Chen,~\IEEEmembership{Member, ~IEEE}, Wei Yuan, Congyan Chen, Shihua Li,~\IEEEmembership{Fellow, ~IEEE}, Hongzhi Yin,~\IEEEmembership{Senior Member, ~IEEE},
\thanks{Manuscript created November, 2025. (Corresponding author: Congyan Chen, Hongzhi Yin)}
\thanks{Dezheng Wang, Congyan Chen, and Shihua Li are with the School of Automation, Southeast University, 210096, China. (e-mail: wangdezheng@seu.edu.cn; chency@seu.edu.cn; lsh@seu.edu.cn)}
\thanks{Tong Chen, Wei Yuan, and Hongzhi Yin are with the School of Electrical Engineering and Computer Science, The University of Queensland, QLD 4072, Australia. (e-mail: tong.chen@uq.edu.au; w.yuan@uq.edu.au; h.yin1@uq.edu.au)}
}



\maketitle

\begin{abstract}
Transformer architectures have shown strong potential in time series forecasting, where multi-head self-attention is widely used to capture temporal dependencies across historical timestamps. However, standard self-attention has quadratic time and memory complexity with respect to the look-back length. This cost may limit its use in resource-constrained or high-throughput forecasting systems, where fast and memory-efficient inference is important. Through qualitative and quantitative analyses, we observe that self-attention maps in time series forecasting often contain redundant patterns across different timestamps. This phenomenon can be related to the repeated temporal patterns and relatively stable temporal correlations in many real-world time series. Motivated by this observation, we propose \textbf{Self-Gating Attention (SGA)}, a plug-and-play attention mechanism that represents the attention score with a shared learnable matrix and an input-dependent residual component. The shared matrix captures common attention patterns, while the residual component captures input-dependent variations. In this way, SGA avoids the query and key projections used in standard attention score computation, leading to linear time and score-matrix memory complexity with respect to the look-back length. We integrate SGA into several forecasting backbones and compare it with standard self-attention and lightweight attention variants on nine publicly available real-world datasets covering electricity, finance, weather, medical monitoring, human activity, and climate records. The results show that SGA improves inference efficiency on public benchmarks while maintaining competitive forecasting performance against state-of-the-art attention mechanisms. These benchmark results provide deployment-oriented evidence.

The code of SGA is available on: \url{https://github.com/DezhengWang/Self-Gating-Attention.git}.
\end{abstract}


\section{Introduction}\label{introduction}
\IEEEPARstart{N}{umerous} time series forecasting models \cite{han_capacity_2024, jin_survey_2024} have been developed for important domains such as weather \cite{ji_spatio_temporal_2024}, finance \cite{bao_long_2025}, industrial applications \cite{wang_timemixer_2024, zhang_multi_task_2025}, and other practical scenarios \cite{wang_investigating_2025}. In industrial settings, efficient time series forecasting can support planning tasks \cite{zhou_informer_2021}, predictive maintenance, and early fault detection, which can help reduce downtime and costs \cite{liu_simultaneous_2019}. Driven by the success of Transformer-based models in natural language processing \cite{vaswani2017attention}, many state-of-the-art time series forecasting models are developed based on Transformer architectures for sequential data modeling. In these models, attention mechanisms, such as standard self-attention (SA) \cite{vaswani2017attention} and its variants, play an important role. By focusing on important historical timestamps and down-weighting less relevant ones, SA captures temporal dependencies in the input sequence. Representative examples include TimeXer \cite{wang_timexer_2024}, SimpleTM \cite{chen_simpletm_2025}, iTransformer \cite{liu_itransformer_2024}, and Informer \cite{zhou_informer_2021}.

\begin{figure}[!t]
\centering
\includegraphics[width=1\columnwidth]{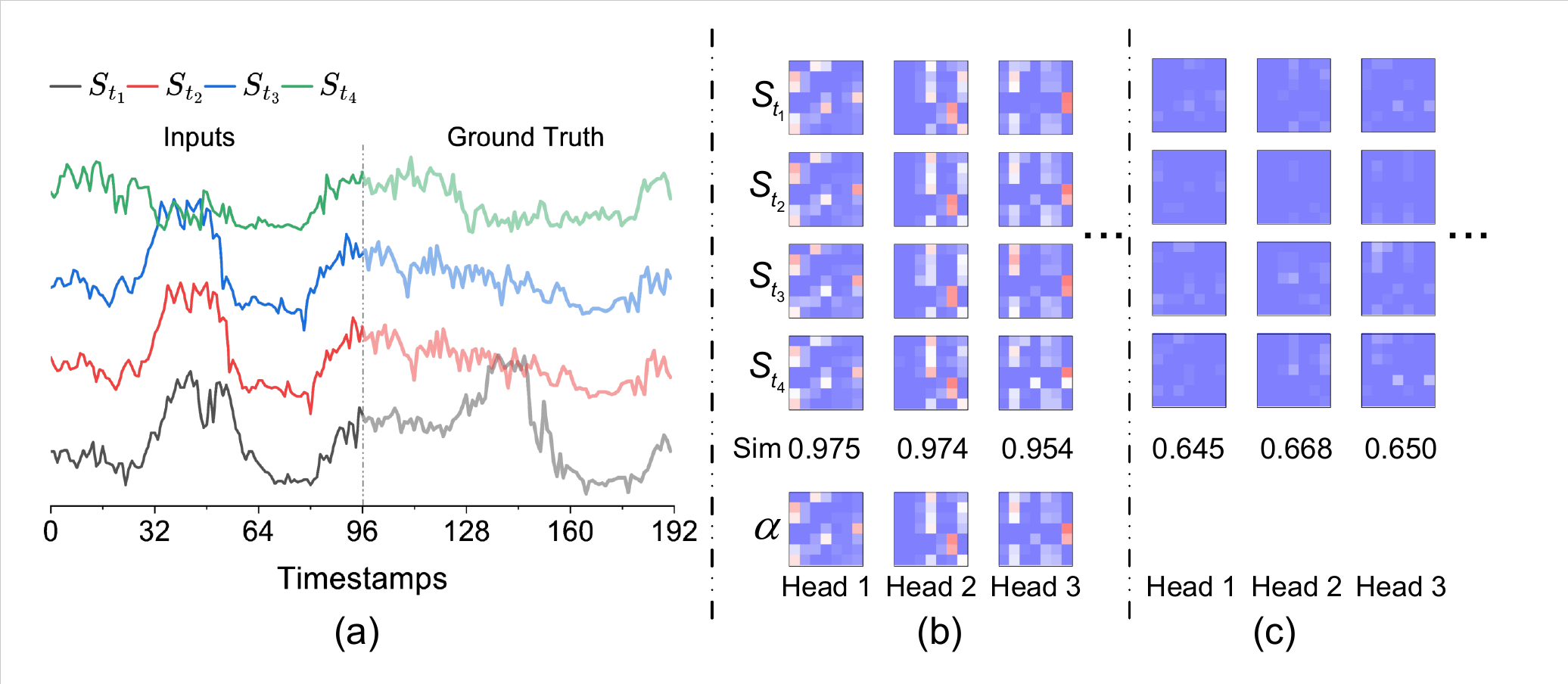}
\caption{
Visualization of attention redundancy on the ETTm1. 
(a) Input trends and ground truth at four selected timestamps. 
(b) SA score maps of the TimeXer backbone. 
(c) Residual score maps obtained by subtracting the mean score map $\alpha$. 
\textit{Sim} denotes the average pairwise cosine similarity.
}
\label{fig123}
\vspace{-0.5cm}
\end{figure}

In standard SA, the input is projected into query, key, and value matrices. The query and key matrices are used to compute attention scores between timestamps, and the value matrix is aggregated according to the normalized attention scores to generate output representations. This process has time and memory complexity that is quadratic with respect to the look-back length, namely the look-back length in time series forecasting \cite{zhou_informer_2021}. As a result, SA becomes an efficiency bottleneck when Transformer-based forecasting models are used in resource-constrained or high-throughput forecasting systems, such as edge devices and traffic-sensor networks \cite{shen_efficient_2021}. This motivates us to examine whether all SA score computations are necessary for time series forecasting. We focus on the following question. \textit{Despite their strong performance, do existing SA mechanisms contain redundant attention score computations?}

To answer this question, we visualize the first-layer SA score maps of TimeXer \cite{wang_timexer_2024} on the ETTm1 \cite{zhou_informer_2021}, as shown in Fig.~\ref{fig123}. ETTm1 is a publicly available real-world benchmark collected from electricity transformer operation in power grid scenarios. We use it as a motivating example to study attention redundancy. In Fig.~\ref{fig123}, $\bm{S}_t$ denotes the attention score matrix, and $\alpha$ denotes the mean of $\bm{S}_{t_1}$, $\bm{S}_{t_2}$, $\bm{S}_{t_3}$, and $\bm{S}_{t_4}$. The residual score maps in Fig.~\ref{fig123}(c) are obtained by subtracting $\alpha$ from the corresponding SA score maps. We select four timestamps, $t_1 = 98$, $t_2 = 964$, $t_3 = 966$, and $t_4 = 2212$. As shown in Fig.~\ref{fig123}(a), the selected timestamps contain both similar and different input patterns. However, the corresponding SA score maps in Fig.~\ref{fig123}(b) still show highly similar patterns across these timestamps. To quantify this similarity, we calculate the average cosine similarity over all timestamp pairs. The similarity values range from 0.885 to 0.975 across attention heads. Based on this observation, we conduct a preliminary experiment by replacing the SA score matrix in each attention head with a learnable shared score matrix. Under this setting, TimeXer obtains an MSE of 0.318, which is close to the original result of 0.315. This suggests that, in this case, the temporal correlations within the input horizon are relatively stable, even when local trends or future values differ. Similar observations on other datasets are reported in Section~\ref{methods}. These results indicate that repeatedly computing SA scores may introduce redundant computation in time series forecasting. This observation is also related to the characteristics of time series data. In natural language processing, different tokens often require strongly context-dependent attention patterns \cite{wu_adversarial_2023}. In contrast, many time series forecasting datasets contain repeated temporal patterns and relatively stable temporal correlations within the input horizon. This suggests that part of the attention score computation can be shared across timestamps, while input-dependent variations should still be preserved. Motivated by this view, we ask the following question: \textit{Can we design a lightweight attention mechanism that captures shared temporal patterns while retaining input-dependent variations?}

\begin{figure*}[!t]
\centering 
\includegraphics[width=2\columnwidth]{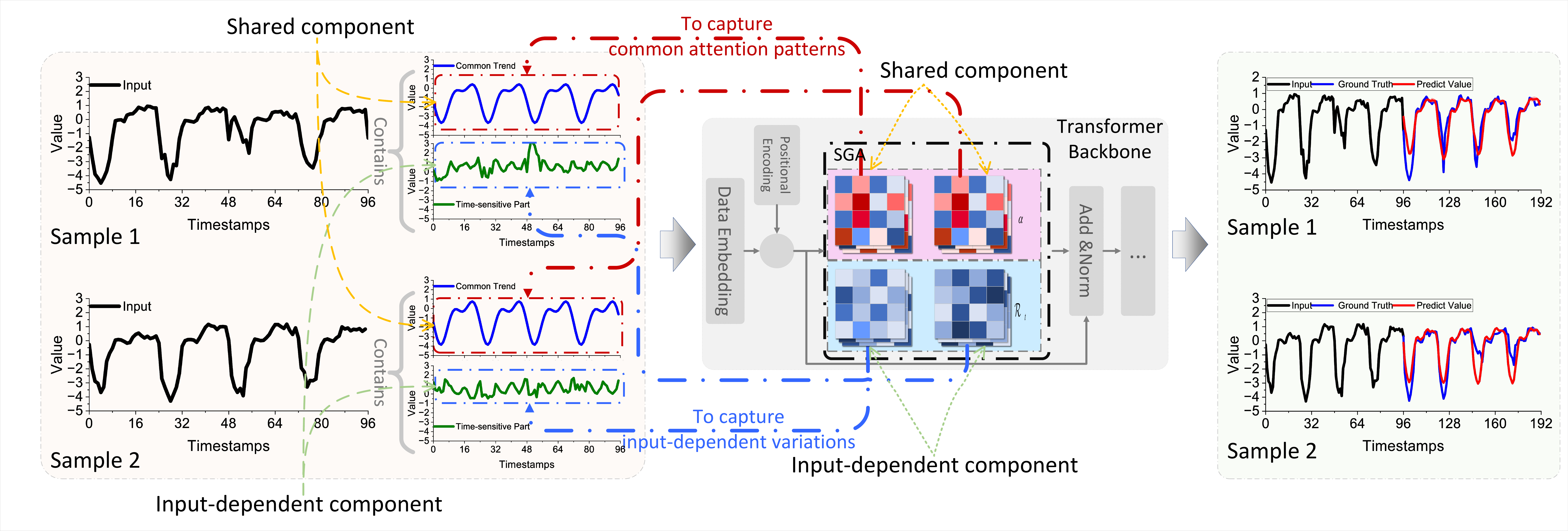} 
\caption{
Illustration of SGA. SGA uses a shared attention score matrix $\bm{\mathcal{A}}$ to model common attention patterns and an input-dependent residual score matrix $\bm{\mathcal{R}}_t$ to capture input-dependent variations.
}
\label{fig_overview}
\end{figure*}

Based on Fig.~\ref{fig123}(b), we further analyze the residual score maps in Fig.~\ref{fig123}(c). In each attention head, the residual score map is obtained by subtracting the mean score map $\alpha$ from the original SA score map. Although these residual score maps have smaller magnitudes, their pairwise similarity is much lower than that in Fig.~\ref{fig123}(b), with the lowest value reaching 0.645. This result suggests that, besides the shared patterns, there still exist input-dependent variations in time series attention. Motivated by this observation, we propose \textbf{Self-Gating Attention (SGA)}. In SGA, the attention score is reparameterized as the combination of a shared attention score matrix and an input-dependent residual score matrix. The shared matrix models common attention patterns across timestamps, while the residual matrix captures input-dependent variations. An illustration of SGA is shown in Fig.~\ref{fig_overview}.

Inspired by TSSA \cite{wu_token_2024}, we use second-order statistics of the input values to construct the residual score matrix, which keeps the residual branch lightweight and avoids extra query-key projections. Different from standard SA and other efficient variants \cite{zhou_informer_2021, wang_long_2024, li_mc_ann_2025}, SGA does not construct additional query and key projections. Instead, its attention score is generated directly from the shared matrix and the input-dependent residual matrix, which motivates the name \textit{self-gating}. In this way, SGA removes two of the three projection operations in standard SA and achieves linear time and score-matrix memory complexity with respect to the look-back length, while maintaining competitive forecasting performance. This paper is motivated by efficient long-sequence forecasting in real-world time series systems, especially monitoring scenarios that require fast and memory-efficient inference. Typical examples include electricity transformer monitoring, edge-device forecasting, and high-throughput sensor analysis. In these settings, models often need to process long historical windows under limited memory or latency constraints. Since standard SA repeatedly computes query-key attention scores, it becomes a major efficiency bottleneck. This motivates us to design a more efficient attention mechanism for time series forecasting. Our main contributions are as follows:
\begin{itemize}
\item We observe that attention score patterns in time series forecasting can be highly similar across different timestamps. A preliminary experiment further shows that a shared attention score matrix can achieve performance close to standard SA, suggesting that repeated input-dependent attention score computation may introduce redundancy.

\item We propose SGA as an efficient alternative to standard SA. SGA replaces query-key attention score computation with a shared attention score matrix and an input-dependent residual score matrix. This design removes the query and key projections used in standard SA and achieves linear time and score-matrix memory complexity with respect to the look-back length.

\item SGA is a model-agnostic attention module that can be plugged into different Transformer-based forecasting backbones. Experiments on multiple real-world time series benchmarks show that SGA improves attention efficiency while maintaining competitive forecasting performance.
\end{itemize}

The rest of this work is organized as follows. Section~\ref{relatedwork} reviews related work. Section~\ref{methods} introduces the proposed method. Section~\ref{results} reports the experimental results. Section~\ref{conclusion} concludes this work.

\section{Related Work}\label{relatedwork}
The Transformer-based models leverage SA, which enables direct modeling of relationships between distant timestamps, thus reducing the maximum information path length to $O(1)$ \cite{vaswani2017attention}. This makes Transformer-based models particularly effective for long sequence time series forecasting \cite{zhou_informer_2021, shibo_hdt_2025} and also widely used for short-term forecasting \cite{eseye_short_term_2020}. TimeXer \cite{wang_timexer_2024} distinguishes and integrates exogenous variables via variate-level cross-attention and endogenous variables via patch-wise SA.  
iTransformer \cite{liu_itransformer_2024} introduces a variable-wise modeling framework. t-PatchGNN \cite{zhang_irregular_2024} transforms each irregular series into time-aligned variable-length patches and applies a time-adaptive graph neural network to capture dependencies, achieving state-of-the-art irregular multivariate time series forecasting performance on healthcare, biomechanics, and climate aspects. UniMATS \cite{ge_unimats_2025} introduces a unified multi-dimensional attention mechanism that jointly attends over space, time, and covariates. TRA-LSTM \cite{pan_temporal_2025} proposes a re-attention loss that corrects and stabilizes temporal attention, aligning it with true process delays. DMA-Trans \cite{yang_difference_2024} contributes an attention mechanism that measures similarity via query–key differences and explicitly encodes positional distance. Although these state-of-the-art models have made significant contributions in the field of time series forecasting, most of them are still based on SA. 

Several variants of SA have been proposed to address the challenges of directly applying SA to time series forecasting. Some aim to improve computational efficiency and reduce parameter size. Others focus on capturing richer temporal relationships. Informer \cite{zhou_informer_2021} selectively attends to the top-$K$ dominant queries that contribute most significantly to the attention distribution. As a result, the computational complexity is reduced from $O(L^2)$ to $O(L \log L)$. Autoformer \cite{wu_autoformer_2021} computes correlations between individual timestamps at the sub-series level. Crossformer \cite{zhang_crossformer_2023} and CAST \cite{wang_cast_2025} introduce dual-level attention structure that separates attention into temporal attention and variable attention levels. SimpleTM \cite{chen_simpletm_2025} introduces a geometric product attention, incorporating an inductive bias toward neighborhood relevance and local-global fusion. Zhao et. al. \cite{zhao_probabilistic_2024} propose a hybrid, two-branch attention with gated fusion that jointly learns inter-load coupling and per-load temporal patterns for multienergy load forecasting. These variations of SA have significantly improved its performance or efficiency in the context of time series forecasting. However, these models are still fundamentally based on SA.

Furthermore, many efficient attention methods are proposed in the field of computer vision or natural language processing. TSSA \cite{wu_token_2024} introduces a novel linear attention mechanism derived from a variational reformulation of the $\text{MCR}^2$ objective; however, it lacks the architectural design to accept an external query, making it unsuitable for cross-attention. Efficient Attention \cite{shen_efficient_2021} decouples the softmax operation and reorders matrix multiplications, making attention more scalable in computer vision tasks. 

Despite the success of attention-based models, many efficient attention mechanisms are adapted from designs originally developed for natural language processing or computer vision \cite{lee_multi-criteria_2024}. 
Current attention-based time series forecasting models \cite{wang_timexer_2024, chen_simpletm_2025, yuan_hierarchical_2024} mainly adopt SA or its variants, but these mechanisms may not fully exploit the recurring patterns in time series data. 
As a result, designing an attention mechanism that is computationally efficient and aligned with time series characteristics remains an important research challenge.

\section{Methods}\label{methods}

\subsection{Analysis of Attention Redundancy}
\label{analysis_redundancy}

\begin{figure}[t]
\centering
\includegraphics[width=1\columnwidth]{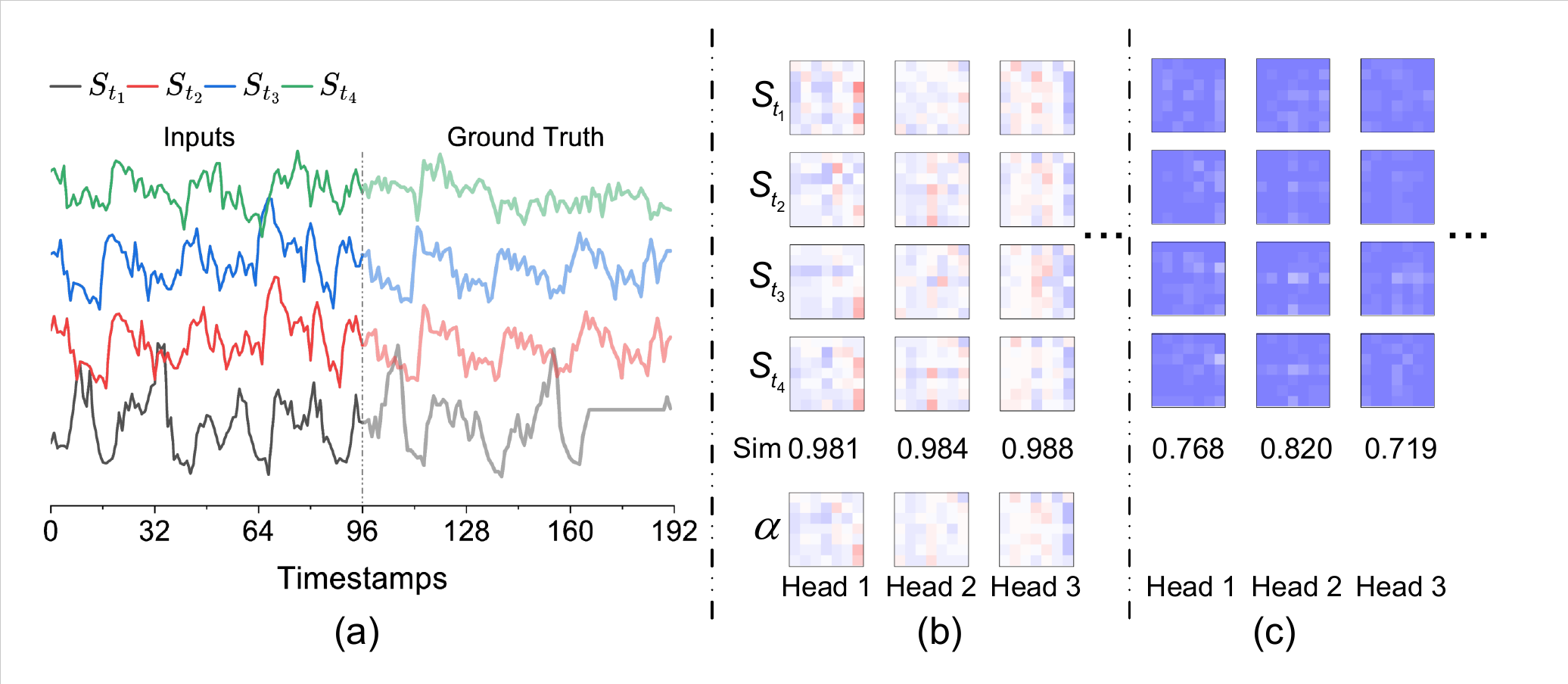}
\caption{
Visualization results on the ETTh1. 
(a) Input trends. 
(b) SA score maps of the TimeXer backbone. 
(c) Residual score maps.
}
\label{timexer_ETTh1}
\end{figure}

\begin{figure}[!t]
\centering
\includegraphics[width=1\columnwidth]{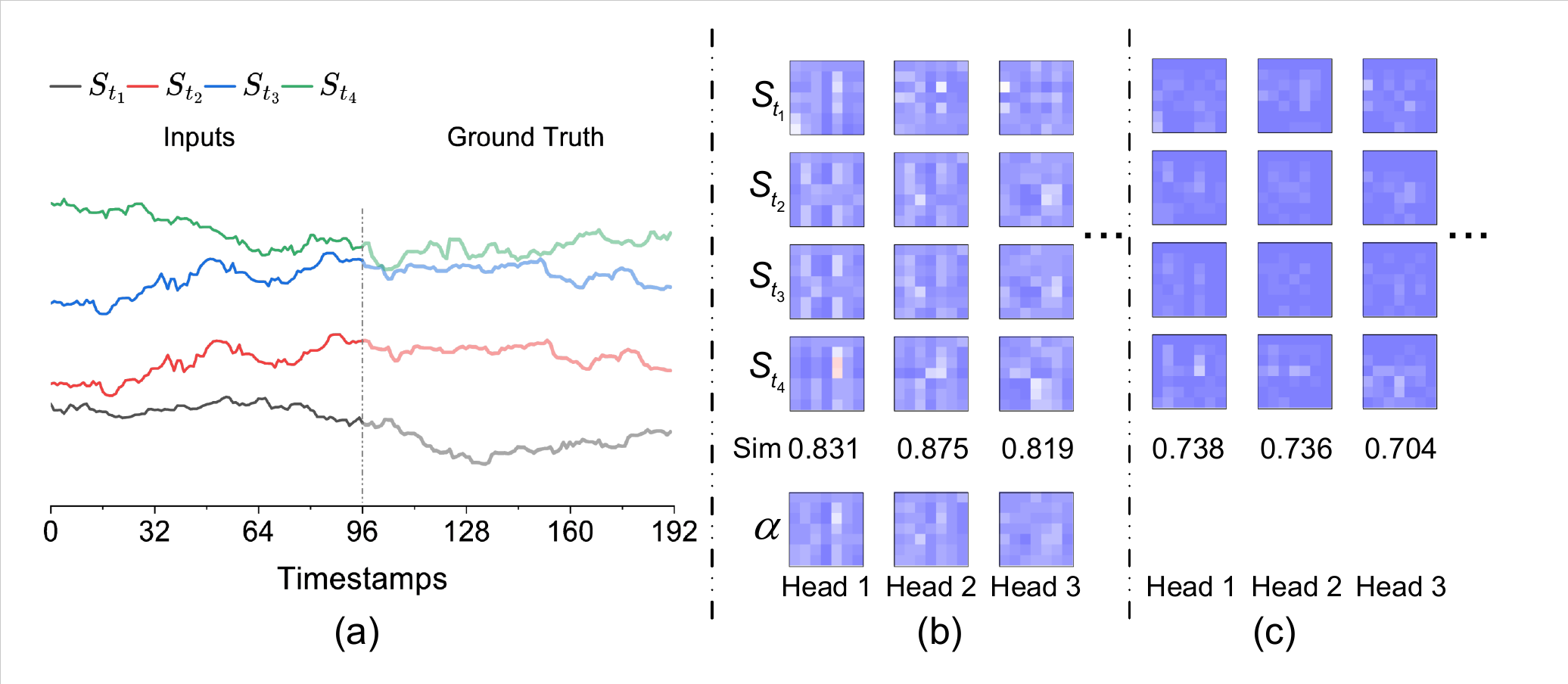}
\caption{
Visualization results on the Exchange-Rate. 
(a) Input trends. 
(b) SA score maps of the TimeXer backbone. 
(c) Residual score maps.
}
\label{timexer_exchange}
\end{figure}

To further examine the attention redundancy observed in Fig.~\ref{fig123}, we visualize the SA score maps on the ETTh1 and Exchange-Rate, as shown in Figs.~\ref{timexer_ETTh1} and \ref{timexer_exchange}. 
We use the same selected timestamps as in Fig.~\ref{fig123} when possible, and set $t_4=1200$ for Exchange-Rate due to its shorter test set. 
These two figures provide additional evidence that the redundancy observed in Fig.~\ref{fig123} is not limited to ETTm1. As shown in Figs.~\ref{timexer_ETTh1} and \ref{timexer_exchange}, patterns similar to those in Fig.~\ref{fig123} can be observed. 
Within each attention head, the SA score maps at different timestamps are highly similar. 
After subtracting the mean score map $\alpha$, the residual score maps show more input-dependent variations than the original SA score maps. 

\begin{figure}[t]
\centering
\includegraphics[width=0.45\columnwidth]{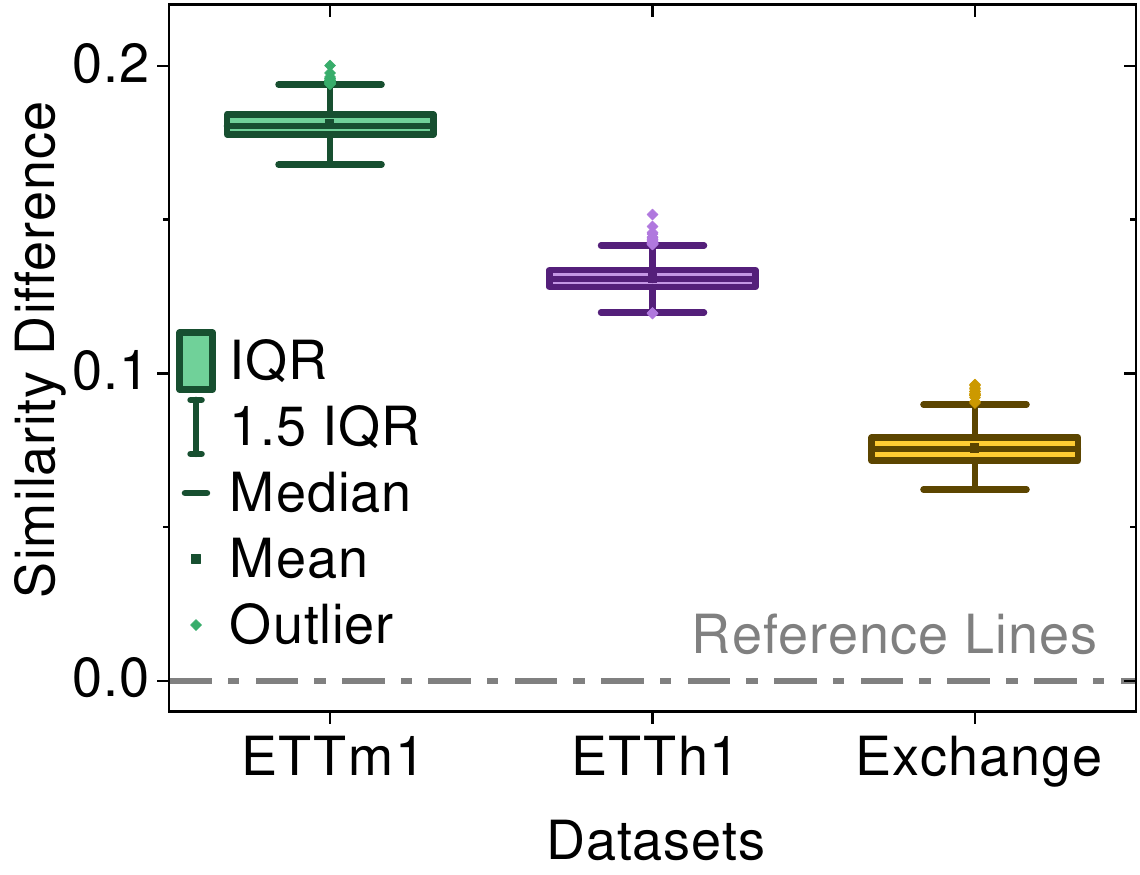}
\llap{(a)~}
\includegraphics[width=0.45\columnwidth]{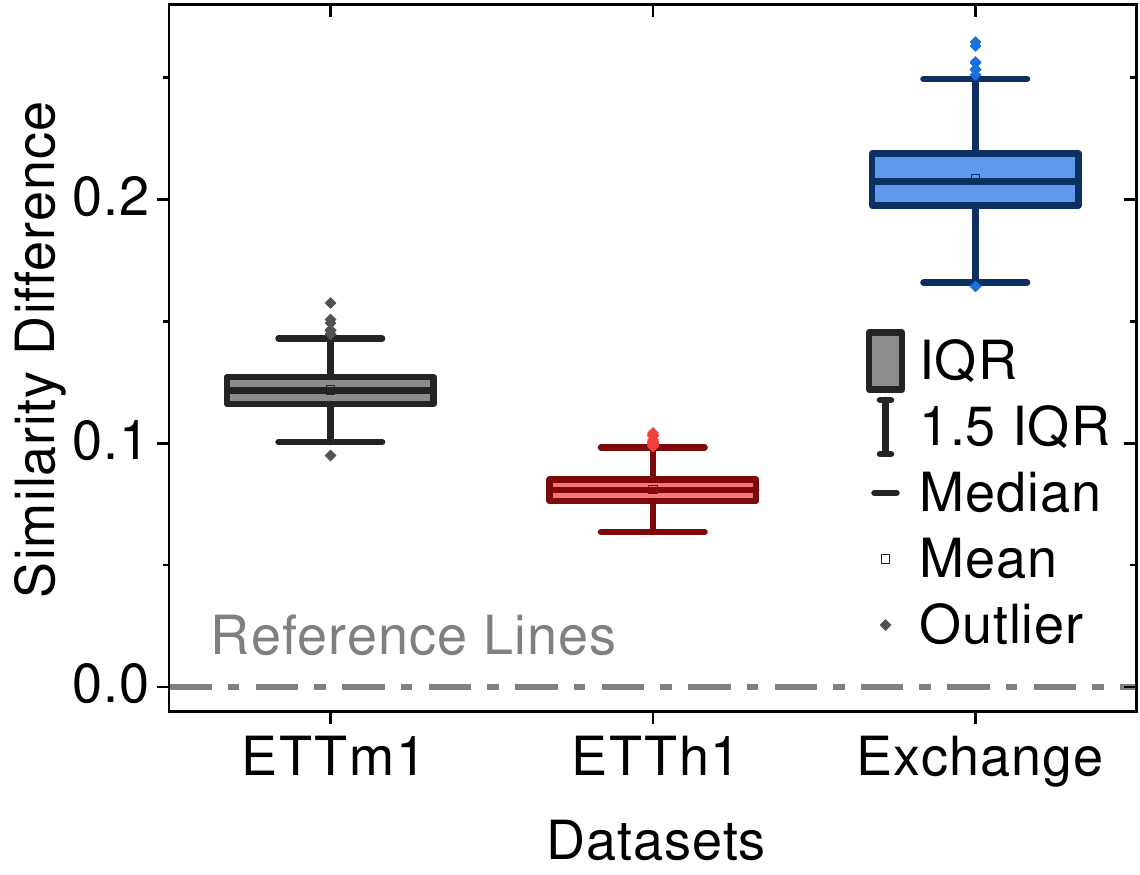}
\llap{(b)~}
\caption{Bootstrap differences between cross-time attention-map similarity and the row-wise shuffled baseline. (a) and (b) report the results for the TimeXer and SimpleTM backbones, respectively.}
\label{fig_bootstrap_redundancy}
\end{figure}

To complement the visual analysis, we conduct a timestamp-level bootstrap test. For each backbone and dataset, we randomly select 100 timestamps from the test set without replacement and extract attention maps from the first encoder layer. We compute cross-time similarity under the same head and compare it with a row-wise shuffled baseline, which preserves row-wise value distributions but removes key-position-specific structures. During bootstrap testing, we resample the selected timestamps with replacement and recompute the similarity difference. For each bootstrap sample, we compute all pairwise cross-time similarities among the resampled timestamps under each head, and then average the results over heads and timestamp pairs. As shown in Fig.~\ref{fig_bootstrap_redundancy} and Table~\ref{tab_bootstrap_similarity}, the differences are consistently positive, and their 95\% confidence intervals are above zero with $p<0.001$. These results suggest that attention maps contain shared structures across timestamps, while the remaining variations are consistent with the use of input-dependent residual score maps.

\begin{table}[t]
\caption{Bootstrap test results for cross-time attention-map similarity. The difference is computed as the observed cross-time similarity minus the row-wise shuffled baseline.}
\centering
\resizebox{\columnwidth}{!}{
\begin{tabular}{ccccc}
\hline
Backbone & Dataset & Difference & 95\% CI of Difference & $p$-value \\
\hline
\multirow{3}{*}{TimeXer} 
& ETTh1    & 0.131 & [0.123, 0.141] & $<0.001$ \\
& ETTm1    & 0.181 & [0.173, 0.192] & $<0.001$ \\
& Exchange & 0.076 & [0.066, 0.088] & $<0.001$ \\
\multirow{3}{*}{SimpleTM} 
& ETTh1    & 0.081 & [0.069, 0.096] & $<0.001$ \\
& ETTm1    & 0.122 & [0.107, 0.139] & $<0.001$ \\
& Exchange & 0.209 & [0.176, 0.243] & $<0.001$ \\
\hline
\end{tabular}}
\label{tab_bootstrap_similarity}
\end{table}

Taken together, the visual and bootstrap results indicate that SA score maps contain both common patterns and input-dependent variations. This observation suggests that repeated input-dependent SA score computation may introduce redundant computation in time series forecasting. Based on this view, SGA represents attention computation with a shared score matrix and an input-dependent residual score matrix. The shared score matrix captures stable patterns across timestamps, while the residual score matrix adjusts the aggregation according to the current input window.

This design may also be applicable to non-stationary sequences, since local input windows may still contain common attention patterns and input-dependent variations. However, when abrupt regime changes or strong concept drift occur, the similarity among SA score maps may become weaker. Stationarization techniques can reduce scale-related distribution shifts \cite{kim_reversible_2021, liu_non-stationary_2022}, making input windows more comparable and shared attention patterns more reasonable.

\subsection{Reparameterizing the Attention Score Matrix}

At timestamp $t$, given an input sequence of $n$ feature vectors 
$\bm{X}_t = [\bm{x}_{t-(n-1)}, \dots, \bm{x}_t]^\mathrm{T} \in \mathbb{R}^{n \times d}$, 
where $d$ is the input dimension, the target sequence is defined as 
$\bm{Y}_t = [\bm{x}_{t+1}, \dots, \bm{x}_{t+s}]^\mathrm{T} \in \mathbb{R}^{s \times d}$, 
where $s$ is the prediction length. 
Before entering the attention unit, a linear transformation 
$f$ \cite{vaswani2017attention, zhou_informer_2021} is applied to $\bm{X}_t$:
\begin{equation}
\bm{V}_t = [\bm{v}_{1,t}, \dots, \bm{v}_{n,t}]^\mathrm{T} = f(\bm{X}_t) \in \mathbb{R}^{n \times \tilde{d}} .
\end{equation}
For simplicity, we omit dimensional changes in the following and write $\tilde{d}=d$. Given the representation matrix $\bm{V}_t$, an attention module aggregates historical representations through a score matrix $\bm{S}_t\in\mathbb{R}^{s\times n}$. The output representation is computed as,
\begin{equation}
\hat{\bm{Y}}_t = \bm{S}_t\bm{V}_t.
\end{equation}

Motivated by the above visual and statistical analyses, we reparameterize the attention score matrix as:
\begin{equation}
\bm{\mathcal{S}}_t= \psi(\bm{\mathcal{A}}, \bm{\mathcal{R}}_t),
\label{eq_def}
\end{equation}
where $\bm{\mathcal{A}} \in\mathbb{R}^{s\times n}$ is a shared attention score matrix across timestamps, and 
$\bm{\mathcal{R}}_t \in\mathbb{R}^{s\times n}$ is an input-dependent residual score matrix computed from $\bm{V}_t$. 

This reparameterization can be understood from a simple additive decomposition. For any timestamp $t$ and prediction step $j$, let $\{z^j_{i,t}\}_{i=1}^{n}$ be attention scores. Given timestamp-independent terms $\{a^j_i\}_{i=1}^{n}$, we define the residual term as:
\begin{equation}
r^j_{i,t}=z^j_{i,t}-a^j_i.
\end{equation}
Then each attention score can be written as:
\begin{equation}
z^j_{i,t}=a^j_i+r^j_{i,t}.
\end{equation}
This form matches the design goal of SGA, where the shared component captures common attention patterns and the residual component captures input-dependent variations.

\subsection{Structure of SGA}

\begin{figure}
\centering
\includegraphics[width=0.8\columnwidth]{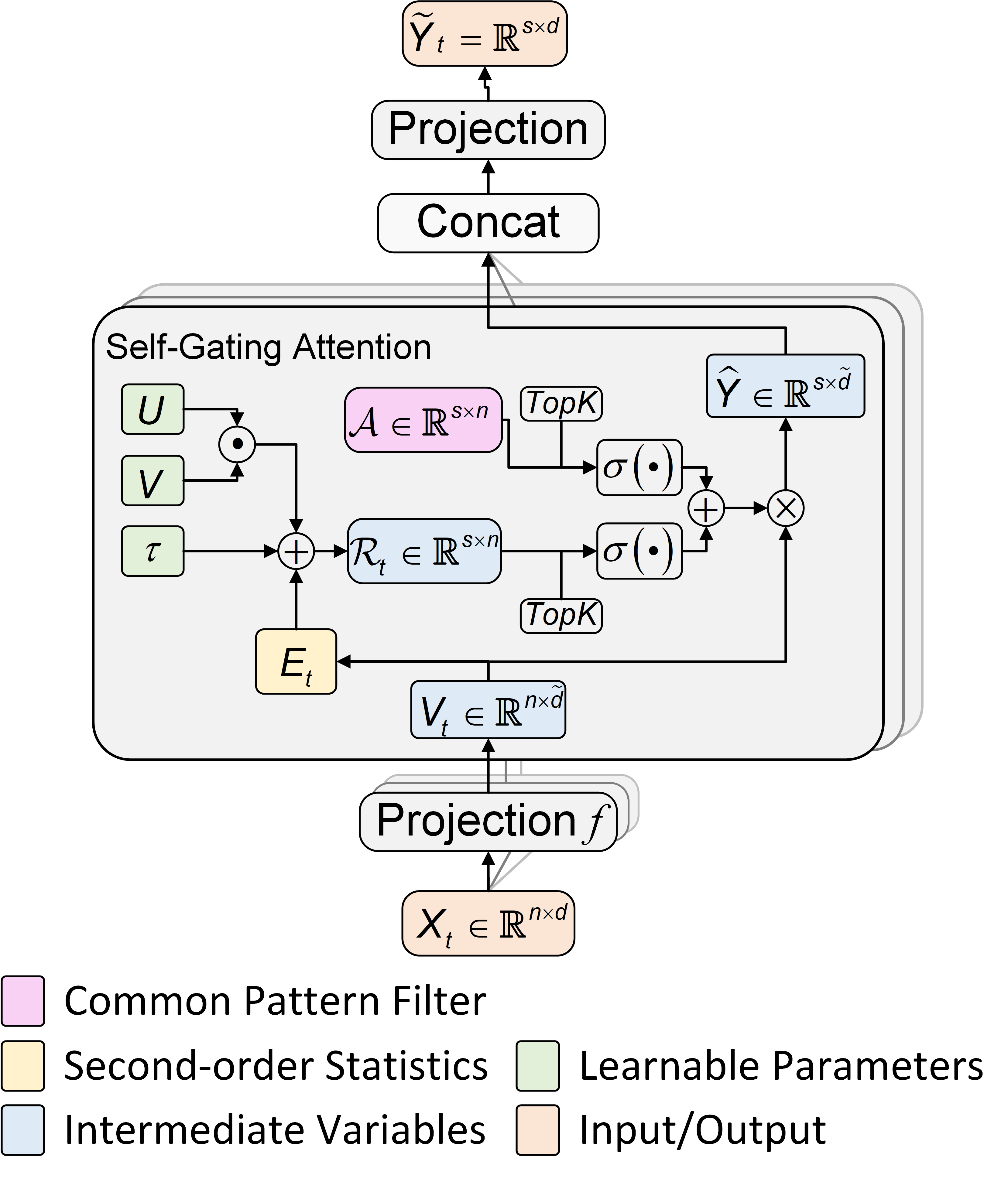}
\caption{Structure of SGA.}
\label{sga}
\end{figure}

SGA is designed to replace the standard SA module in Transformer-based forecasting models. 
Its complexity reduction comes from a different source from existing efficient attention variants. 
Instead of approximating or sparsifying query-key attention scores after they are computed, SGA avoids repeated query-key score computation. 
It uses a shared attention score matrix to model common attention patterns and an input-dependent residual score matrix to capture input-dependent variations. 
The core structure is shown in Fig.~\ref{sga}.

Based on the reparameterization in Eq.~(\ref{eq_def}), the output of SGA at timestamp $t$ is written as:
\begin{equation}
    \bm{\hat{Y}}_t
    =
    \bm{\mathcal{S}}_t\bm{V}_t
    =
    \psi(\bm{\mathcal{A}},\bm{\mathcal{R}}_t)\bm{V}_t,
    \label{eq_sga_def}
\end{equation}
where $\bm{\mathcal{A}}$ is the shared attention score matrix, $\bm{\mathcal{R}}_t$ is the input-dependent residual score matrix, and $\psi(\cdot)$ denotes the fusion operation. 
The shared matrix $\bm{\mathcal{A}}$ can be viewed as a common attention prior rather than the whole attention score matrix. 
It captures stable and repeated patterns learned from training data. 
The residual matrix $\bm{\mathcal{R}}_t$ is computed from the current input window, so it can adjust the attention score when local trends or operating conditions change.

\textbf{Shared attention score matrix $\bm{\mathcal{A}}$.}
The matrix $\bm{\mathcal{A}}\in\mathbb{R}^{s\times n}$ is shared across timestamps and does not change with $t$. 
It is used to capture common attention patterns in time series. 
Compared with recomputing attention scores at every timestamp, the shared matrix reduces repeated computation over similar score patterns. This is especially useful for long input sequences, where repeated attention score computation becomes more expensive.

For multi-head attention, each attention head $h$ has an independent shared score matrix $\bm{\mathcal{A}}^h$. 
To reduce redundancy among attention heads, we use orthogonal initialization:
\begin{equation}
    \left\langle 
    \mathrm{vec}(\bm{\mathcal{A}}^h),
    \mathrm{vec}(\bm{\mathcal{A}}^{h'})
    \right\rangle = 0,
    \quad h\neq h',
    \label{eq_orthogonal_init}
\end{equation}
where $\mathrm{vec}(\cdot)$ denotes vectorization. 
This strategy is only used at initialization. 
It does not add an extra loss term and does not change the training objective.

\textbf{Input-dependent residual score matrix $\bm{\mathcal{R}}_t$.}
The shared matrix $\bm{\mathcal{A}}$ captures common attention patterns, but it may not fully describe local changes in different input windows. 
Therefore, SGA introduces an input-dependent residual score matrix $\bm{\mathcal{R}}_t\in\mathbb{R}^{s\times n}$. 
This matrix changes with $\bm{V}_t$ and helps capture input-dependent variations.

There are many possible ways to construct $\bm{\mathcal{R}}_t$. Since SGA aims to reduce parameters and FLOPs, its construction should preserve useful input-dependent information while keeping the attention computation lightweight. We first examine the query-key score computation in standard self-attention. For simplicity, we omit bias terms. The query and key matrices are computed as $\bm{Q}_t=\bm{X}_t\bm{W}_Q$ and $\bm{K}_t=\bm{X}_t\bm{W}_K$, respectively. Therefore, the score matrix can be written as,
\begin{equation}    
\bm{Q}_t\bm{K}_t^\mathrm{T}
= \bm{X}_t\bm{W}_Q\bm{W}_K^\mathrm{T}\bm{X}_t^\mathrm{T}.
\end{equation}
This form shows that the attention score matrix depends on pairwise second-order interactions of the input representation $\bm{X}_t$. To reduce computational complexity, TSSA~\cite{wu_token_2024} constructs attention-like transformations from token statistics rather than pairwise query-key similarity. More specifically, TSSA uses empirical second-moment statistics of token features, showing that second-order token statistics can provide useful information for attention computation. In addition, several studies~\cite{li_occlusion-aware_2024, wang_tamt_2025} show that second-order information can capture rich feature correlations and support feature recalibration or representation enhancement. Following this statistics-based view, we construct $\bm{\mathcal{R}}_t$ from the normalized second-order statistics of $\bm{V}_t$ to keep the residual branch lightweight. Compared with directly using $\bm{X}_t$, $\bm{V}_t$ is obtained through a learnable linear projection and is also used in the final attention aggregation. Thus, its second-order statistics provide a task-adapted residual cue for adjusting the shared score matrix. Specifically, the second-order statistic summarizes the representation energy at each historical position. A larger energy indicates a stronger response in the current input window, which provides useful information for adjusting the shared attention pattern according to input-dependent changes. This construction is suitable for the residual branch because $\bm{\mathcal{R}}_t$ is only expected to adjust the shared score matrix according to input-dependent changes, rather than reconstruct a full query-key attention map. For each historical position, we define:
\begin{equation}
    e_{i,t}
    =
    \frac{1}{d}
    \sum_{j=1}^{d}
    [\bm{V}_t]_{i,j}^{2},
    \quad i=1,\dots,n.
    \label{eq_energy}
\end{equation}
Let $\bm{e}_t=(e_{1,t},\dots,e_{n,t})^{\mathrm{T}}\in\mathbb{R}^{n\times 1}$. 
The normalized energy is defined as:
\begin{equation}
    \bar{e}_t
    =
    \sqrt{
    \frac{1}{n}
    \sum_{i=1}^{n}
    e_{i,t}
    },
    \quad
    \bm{E}_t
    =
    \frac{\bm{e}_t}{\bar{e}_t}
    \in\mathbb{R}^{n\times 1}.
    \label{eq_normalized_energy}
\end{equation}
Positions with larger normalized energy usually indicate stronger representation responses in the current window, which can help the residual branch adjust the shared attention pattern. The normalization step reduces scale differences across input windows, making the residual score more stable and comparable across samples.

For multi-head attention, let $H$ be the number of heads and let $\bm{\gamma}\in\mathbb{R}^{H}$ be a learnable scaling vector. 
For the $h$-th attention head, the residual score matrix is defined as:
\begin{equation}
    \bm{\mathcal{R}}_t^h
    =
    \mathbf{1}_s
    \left(
    \operatorname{sp}(\gamma_h)\bm{E}_t
    \right)^{\mathrm{T}}
    +
    \bm{\tau}^{h}
    +
    \bm{B}^{h},
    \label{eq_residual}
\end{equation}
where $\mathbf{1}_s\in\mathbb{R}^{s\times 1}$ is an all-one vector, $\operatorname{sp}(\cdot)$ denotes the softplus function, and $\bm{\tau}^{h}\in\mathbb{R}^{s\times n}$ is a learnable bias. 
The term $\bm{B}^{h}$ is a low-rank bilinear term:
\begin{equation}
    \bm{B}^{h}
    =
    \bm{U}^{h}\bm{W}^{h},
    \quad
    \bm{U}^{h}\in\mathbb{R}^{s\times r},
    \quad
    \bm{W}^{h}\in\mathbb{R}^{r\times n}.
    \label{eq_low_rank}
\end{equation}
For simplicity, we omit the head index $h$ in the following and write the residual score matrix as $\bm{\mathcal{R}}_t$.

\textbf{Attention score fusion.}
Before fusing $\bm{\mathcal{A}}$ and $\bm{\mathcal{R}}_t$, we apply top-$K$ sparsification at the logit level. 
This operation keeps the largest $K$ elements in each row and masks the remaining elements with $-\infty$. 
It helps reduce the influence of weakly related positions. 
For the $j$-th row of $\bm{\mathcal{Z}}$, let $\mathcal{I}_K(\bm{\mathcal{Z}}_{j,:})$ denote the index set of the largest $K$ elements. For any score matrix $\bm{\mathcal{Z}}\in\mathbb{R}^{s\times n}$, the row-wise top-$K$ sparsified matrix is defined as:
\begin{equation}
[\operatorname{TopK}(\bm{\mathcal{Z}})]_{j,i}
=
\begin{cases}
\bm{\mathcal{Z}}_{j,i}, & i\in \mathcal{I}_K(\bm{\mathcal{Z}}_{j,:}),\\
-\infty, & \mathrm{otherwise},
\end{cases}
\quad j=1,\dots,s.
\label{eq_topk_sparse}
\end{equation}
After sparsification, the fusion function is defined as:
\begin{equation}
    \psi(\bm{\mathcal{A}},\bm{\mathcal{R}}_t)
    =
    \operatorname{softmax}(\operatorname{TopK}(\bm{\mathcal{A}}))
    +
    \operatorname{softmax}(\operatorname{TopK}(\bm{\mathcal{R}}_t)),
    \label{eq_fusion}
\end{equation}
where the softmax is applied row-wise. 
The final output of SGA is:
\begin{equation}
    \bm{\hat{Y}}_t
    =    \left(\psi(\bm{\mathcal{A}},\bm{\mathcal{R}}_t)
    \right)
    \bm{V}_t.
    \label{eq_sga_final}
\end{equation}

Compared with standard SA, SGA does not construct query and key projections. 
It only keeps the value projection and computes attention scores through a shared attention score matrix and a lightweight input-dependent residual branch. 
For one attention head, the multiplication between the fused score matrix and $\bm{V}_t$ costs $O(snd)$. 
When $d$ is fixed, the cost can be written as $O(sn)$, and the memory cost of the score matrix is $O(sn)$. 
Thus, SGA scales linearly with the look-back length $n$ and the prediction length $s$.

\subsection{Generalizing SGA for Cross-Attention}

The proposed SGA can replace existing attention modules in Transformer-based time series forecasting models. 
In its default form, SGA models the dependencies within the input sequence up to timestamp $t$, similar to standard SA. 
Some forecasting models also use cross-attention to incorporate latent representations of the target sequence or future exogenous variables \cite{kim_are_2024}. 
SGA can be extended to such settings with two minor changes.

First, the input vectors $\bm{X}_t \in \mathbb{R}^{n \times d}$ and the query vectors $\bm{Q}_t \in \mathbb{R}^{s \times d}$ are concatenated along the temporal axis, giving:
\begin{equation}
\tilde{\bm{X}}_t \in \mathbb{R}^{(n+s) \times d}.
\end{equation}
The combined sequence is then passed through the projection layer $f$ to obtain the value matrix $\bm{V}_t$.

Second, the shared attention score matrix is changed to:
\begin{equation}
\bm{\mathcal{A}} \in \mathbb{R}^{s \times (n+s)},
\end{equation}
so that SGA can aggregate information from the combined input sequence into the target sequence length.

This extension allows SGA to be used in both self-attention and cross-attention settings, making it compatible with different Transformer-based forecasting backbones.

\section{Experiments}\label{results}
\subsection{Settings}
\begingroup
\renewcommand{\thefootnote}{}
\footnotetext{\textit{Link updated as of April 26, 2026}:}
\endgroup

\textbf{Datasets:}
We evaluate SGA on both regular and irregular time series forecasting datasets. 
Regular time series use a uniform sampling frequency, while irregular time series contain variables that are asynchronously sampled at different frequencies. 
For regular time series, we use the widely used ETT datasets, including ETTh1, ETTh2, ETTm1, and ETTm2\footnote{https://github.com/zhouhaoyi/ETDataset} \cite{zhou_informer_2021}, Weather\footnote{https://www.bgc-jena.mpg.de/wetter/}, and Exchange-Rate\footnote{https://dl.acm.org/doi/pdf/10.1145/3209978.3210006}. 
These datasets correspond to electricity transformer operation, environmental observation, and financial exchange rates, respectively. 
For irregular time series, we use three benchmarks, including the medical dataset PhysioNet ICU\footnote{https://archive.physionet.org/challenge/2012}, the biomechanics dataset Human Activity\footnote{https://archive.ics.uci.edu/dataset/196/localization+data+for+person+activity}, and the climate dataset USHCN\footnote{https://www.osti.gov/biblio/1394920}. 
These datasets are used to examine whether SGA can generalize beyond regular time series settings. 
Detailed dataset specifications are provided in Table~\ref{dataset}.

All datasets used in our experiments are publicly available real-world datasets released by prior studies.  
Among them, the ETT datasets are used as a representative industrial application case related to electricity transformer forecasting in power grid scenarios. 
This case is relevant to industrial monitoring, where forecasting models often need to process long historical windows while keeping the computation and memory costs low.

\begin{table}[t!]
\caption{Specifications of the datasets. `Freq.' is a shorthand for sampling frequency. ‘$^\ddag$’ denotes irregular datasets and the others are regular.}
\centering
\resizebox{\columnwidth}{!}{
\begin{tabular}{cccc}
\hline
Dataset & Variates & Dataset Size & Freq. \\
\hline
ETTh1, ETTh2 & 7 & (8545, 2881, 2881) & Hourly \\
ETTm1, ETTm2 & 7 & (34465, 11521, 11521) & 15 min \\
Weather & 21 & (36792, 5271, 10540) & 10 min \\
Exchange-Rate & 8 & (5312, 1517, 760) & Daily \\
Human Activity$^\ddag$ & 12 & (3363, 1022, 1089) & ---\\
PhysioNet ICU$^\ddag$ & 41 & (7200, 2400, 2400) & ---\\
USHCN$^\ddag$ & 5 & (668, 223, 223) & ---\\
\hline
\end{tabular}}
\label{dataset}
\end{table}

\textbf{Baselines:} Specifically, we evaluate against standard SA \cite{vaswani2017attention}, Geometry attention \cite{chen_simpletm_2025}, ProbSparse attention \cite{zhou_informer_2021}, AutoCorrelation \cite{wu_autoformer_2021}, and TSSA \cite{wu_token_2024}. For regular time series datasets, all attention mechanisms are paired with seven representative backbone models, namely SimpleTM \cite{chen_simpletm_2025}, iTransformer \cite{liu_itransformer_2024}, TimeXer \cite{wang_timexer_2024}, CARD \cite{wang_card_nodate}, FEDformer \cite{zhou_fedformer_2022}, PAttn \cite{tan_are_2024}, and MultiPatchFormer \cite{vahid_multiscale_2025}. For irregular time series datasets, all attention variants are implemented with the t-PatchGNN \cite{zhang_irregular_2024} backbone, which is a representative backbone for irregular time series forecasting.

\textbf{Setup:}
We use MSE and MAE as evaluation metrics. 
All experiments are repeated three times, and the averaged results are reported. 
For regular time series, all datasets are normalized to have zero mean and unit variance before processing \cite{zhou_informer_2021}. 
We use prediction lengths of 96, 192, 336, and 720, following prior works \cite{zhou_informer_2021, wang_timexer_2024, chen_simpletm_2025}. 
Before each batch, we apply series stationarization to the input sequence \cite{liu_non-stationary_2022}, and convert the outputs back using the input statistics \cite{liu_itransformer_2024}. For each backbone, all attention variants are evaluated under the same backbone training script and hyper-parameter configuration, unless otherwise required by the original attention module.
All experiments are run on a single NVIDIA A100 GPU with 40GB memory. These experiments provide controlled benchmark comparisons, but they do not directly validate deployment performance on resource-constrained industrial devices.

\subsection{Effectiveness Results}
\begin{table*}[!t]
\caption{Time series forecasting results. The table reports MSE and MAE averaged across 4 prediction lengths, along with their standard deviations (SD). The best values are shown in bold and the second-best values are underlined, which are determined using the unrounded results. `\# Top-2' denotes the total number of best and second-best results.
}
\centering
\resizebox{2\columnwidth}{!}{
\begin{tabular}{@{}c@{\hspace{5pt}}c@{\hspace{5pt}}c@{\hspace{5pt}}c@{\hspace{5pt}}c@{\hspace{5pt}}c@{\hspace{5pt}}c@{\hspace{5pt}}c@{\hspace{5pt}}c@{\hspace{5pt}}c@{\hspace{5pt}}c@{\hspace{5pt}}c@{\hspace{5pt}}c@{}@{\hspace{5pt}}c@{}@{\hspace{5pt}}c@{}}
\hline
\multirow{3}{*}{Model} & \multirow{3}{*}{Dataset} & \multicolumn{2}{c}{SGA*} & \multicolumn{2}{c}{SA} & \multicolumn{2}{c}{Geometry} & \multicolumn{2}{c}{ProbSparse} & \multicolumn{2}{c}{AutoCorrelation} & \multicolumn{2}{c}{TSSA}\\
& & MSE & MAE & MSE & MAE & MSE & MAE & MSE & MAE & MSE & MAE & MSE & MAE\\
& & ($10^{-3}$ SD) & ($10^{-3}$ SD) & ($10^{-3}$ SD) & ($10^{-3}$ SD) & ($10^{-3}$ SD) & ($10^{-3}$ SD) & ($10^{-3}$ SD) & ($10^{-3}$ SD) & ($10^{-3}$ SD) & ($10^{-3}$ SD) & ($10^{-3}$ SD) & ($10^{-3}$ SD)\\\hline

\multirow{7}{*}{\rotatebox{90}{TimeXer}}
 & ETTh1 & 
 \textbf{0.435(2.58)} & \underline{0.431(2.06)} & 
 \underline{0.442(5.52)} & 0.437(3.04) & 
 0.450(3.75) & 0.442(3.27) & 
 0.456(4.65) & 0.443(2.54) & 
 0.442(3.84) & \textbf{0.430(1.94)} & 
 0.448(5.54) & 0.434(3.75)\\
 
 & ETTh2 & 
 \textbf{0.366(2.95)} & \textbf{0.393(2.50)} & 
 0.374(4.88) & 0.398(3.40) & 
 0.376(4.04) & 0.400(2.60) & 
 \underline{0.370(3.48)} & \underline{0.396(2.61)} & 
 0.375(5.97) & 0.399(4.81) & 
 0.383(15.17) & 0.402(7.75)\\
 
 & ETTm1 & 
 \textbf{0.377(1.22)} & \textbf{0.388(1.09)} & 
 0.381(1.57) & 0.392(1.14) & \underline{0.379(2.16)} & 
 0.390(1.67) & 0.392(2.02) & 0.399(1.53) & 
 0.379(2.51) & \underline{0.389(1.51)} & 
 0.385(2.54) & 0.393(2.21)\\
 
 & ETTm2 & 
 \textbf{0.271(1.28)} & \textbf{0.317(1.03)} & 
 0.274(1.45) & \underline{0.319(0.64)} & 
 \underline{0.274(1.45)} & 0.320(1.30) & 
 0.275(3.01) & 0.319(1.72) & 
 0.274(1.03) & 0.320(1.01) & 
 0.276(1.90) & 0.322(1.47)\\
 
 & Exchange & 
 \textbf{0.363(6.55)} & \textbf{0.404(3.57)} & 
 0.377(10.99) & 0.411(6.17) & 
 0.377(10.75) & 0.412(5.78) & 
 \underline{0.372(2.82)} & \underline{0.410(1.52)} & 
 0.380(6.74) & 0.415(4.03) & 
 0.382(2.96) & 0.415(1.85)\\
 
 & Weather & 
 \textbf{0.237(0.52)} & \textbf{0.264(0.42)} & 
 0.241(1.21) & 0.267(0.97) & 
 0.240(0.62) & 0.266(0.63) & 
 0.241(0.87) & 0.267(0.72) & 
 0.242(0.74) & 0.268(0.77) & 
 \underline{0.240(1.69)} & \underline{0.266(0.86)}\\\hline
 
\multicolumn{2}{c}{\# Top-2}& 
\multicolumn{2}{c}{\textbf{12}} &
\multicolumn{2}{c}{2} &
\multicolumn{2}{c}{2} &
\multicolumn{2}{c}{4} &
\multicolumn{2}{c}{2} &
\multicolumn{2}{c}{2} \\\hline

\multirow{7}{*}{\rotatebox{90}{iTransformer}}
 & ETTh1 &
 \underline{0.450(1.11)} & \underline{0.441(1.58)} &
 0.452(4.73) & 0.443(3.23) &
 0.467(3.58) & 0.452(2.66) &
 0.476(6.15) & 0.456(3.89) &
 0.456(5.35) & 0.444(3.89) &
 \textbf{0.450(4.29)} & \textbf{0.439(1.85)}\\
 
 & ETTh2 &
 0.380(1.13) & 0.404(0.65) &
 0.379(1.44) & 0.402(0.87) &
 0.383(3.40) & 0.404(1.76) &
 0.379(5.56) & 0.403(2.84) &
 \textbf{0.376(2.41)} & \textbf{0.401(2.04)} &
 \underline{0.378(3.24)} & \underline{0.401(2.29)}\\
 
 & ETTm1 &
 \textbf{0.391(1.99)} & \underline{0.397(1.39)} &
 0.401(1.52) & 0.402(0.67) &
 0.403(2.06) & 0.404(1.17) &
 0.395(1.18) & 0.399(0.68) &
 0.407(2.81) & 0.404(2.38) &
 \underline{0.392(2.47)} & \textbf{0.397(1.22)}\\
 
 & ETTm2 &
 \underline{0.283(0.55)} & 0.326(0.44) &
 0.286(1.00) & 0.329(0.95) &
 0.287(1.11) & 0.329(1.06) &
 \textbf{0.281(1.48)} & \textbf{0.324(1.09)} &
 0.283(1.32) & 0.326(0.96) &
 0.283(1.51) & \underline{0.325(1.35)}\\
 
 & Exchange &
 \textbf{0.361(1.69)} & \textbf{0.406(0.97)} &
 0.367(0.95) & 0.408(0.60) &
 \underline{0.365(2.90)} & 0.408(2.02) &
 0.374(5.48) & 0.413(3.33) &
 0.372(4.75) & 0.410(2.54) &
 0.366(1.44) & \underline{0.407(0.78)}\\
 
 & Weather &
 \underline{0.244(0.86)} & \underline{0.270(1.20)} &
 0.258(2.35) & 0.277(1.50) &
 0.258(1.51) & 0.277(1.24) &
 \textbf{0.244(0.96)} & \textbf{0.269(1.04)} &
 0.260(2.57) & 0.279(2.30) &
 0.246(2.05) & 0.271(1.46)\\\hline
 
\multicolumn{2}{c}{\# Top-2}&
\multicolumn{2}{c}{\textbf{9}} &
\multicolumn{2}{c}{0} &
\multicolumn{2}{c}{1} &
\multicolumn{2}{c}{4} &
\multicolumn{2}{c}{2} &
\multicolumn{2}{c}{8} \\\hline

\multirow{7}{*}{\rotatebox{90}{SimpleTM}}
 & ETTh1 &
 \textbf{0.434(1.42)} & \textbf{0.431(1.13)} &
 0.442(4.56) & 0.438(3.15) &
 0.437(2.52) & \underline{0.434(2.44)} &
 \underline{0.437(2.21)} & 0.435(1.39) &
 0.447(6.13) & 0.440(3.77) &
 0.441(2.61) & 0.436(1.77)\\
 
 & ETTh2 &
 \textbf{0.371(1.48)} & \textbf{0.398(0.76)} &
 0.376(3.31) & 0.400(1.48) &
 \underline{0.374(4.48)} & \underline{0.398(2.70)} &
 0.377(1.75) & 0.400(0.67) &
 0.382(6.24) & 0.404(3.77) &
 0.377(4.17) & 0.401(2.17)\\
 
 & ETTm1 &
 \textbf{0.379(0.39)} & \textbf{0.388(0.35)} &
 0.383(1.88) & 0.390(1.14) &
 0.381(0.66) & 0.390(0.83) &
 \underline{0.381(0.31)} & \underline{0.389(0.29)} &
 0.391(3.68) & 0.396(3.59) &
 0.389(4.65) & 0.395(2.84)\\
 
 & ETTm2 &
 \textbf{0.273(0.53)} & \textbf{0.318(0.70)} &
 0.276(0.75) & 0.320(0.66) &
 \underline{0.274(1.42)} & \underline{0.319(1.12)} &
 0.275(1.61) & 0.320(1.07) &
 0.279(2.11) & 0.324(2.12) &
 0.276(1.85) & 0.320(1.71)\\
 
 & Exchange &
 \textbf{0.375(2.71)} & \textbf{0.413(1.50)} &
 0.389(13.40) & 0.422(6.77) &
 \underline{0.383(8.45)} & \underline{0.417(5.30)} &
 0.401(17.36) & 0.428(10.79) &
 0.396(7.40) & 0.424(4.42) &
 0.392(12.04) & 0.422(8.11)\\
 
 & Weather &
 \underline{0.245(0.12)} & \underline{0.269(0.21)} &
 0.245(0.60) & 0.270(0.71) &
 \textbf{0.241(1.12)} & \textbf{0.266(1.09)} &
 0.246(0.70) & 0.270(0.67) &
 0.270(2.55) & 0.289(1.91) &
 0.250(1.76) & 0.276(2.09)\\\hline
 
\multicolumn{2}{c}{\# Top-2}&
\multicolumn{2}{c}{\textbf{12}} &
\multicolumn{2}{c}{0} &
\multicolumn{2}{c}{9} &
\multicolumn{2}{c}{3} &
\multicolumn{2}{c}{0} &
\multicolumn{2}{c}{0} \\\hline

\multirow{7}{*}{\rotatebox{90}{CARD}}
 & ETTh1 &
 \underline{0.445(0.46)} & \underline{0.432(0.58)} &
 0.450(4.43) & 0.438(2.88) &
 0.448(1.18) & 0.435(1.84) &
 0.447(2.40) & 0.433(2.49) &
 0.447(1.30) & 0.435(0.79) &
 \textbf{0.444(1.54)} & \textbf{0.431(0.93)}\\
 
 & ETTh2 &
 \textbf{0.364(2.28)} & \textbf{0.393(1.23)} &
 \underline{0.370(2.92)} & \underline{0.395(1.22)} &
 0.370(1.93) & 0.395(1.38) &
 0.374(4.12) & 0.397(1.91) &
 0.376(1.61) & 0.397(0.87) &
 0.381(1.36) & 0.400(0.84)\\
 
 & ETTm1 &
 \underline{0.392(0.87)} & 0.396(0.86) &
 \textbf{0.387(3.04)} & \textbf{0.394(1.56)} &
 0.397(1.54) & \underline{0.395(1.41)} &
 0.394(1.89) & 0.396(0.75) &
 0.404(1.52) & 0.398(1.76) &
 0.394(2.91) & 0.395(1.16)\\
 
 & ETTm2 &
 \underline{0.273(1.49)} & \underline{0.319(0.61)} &
 \textbf{0.271(3.91)} & \textbf{0.318(2.71)} &
 0.280(1.46) & 0.323(0.60) &
 0.280(2.81) & 0.323(1.86) &
 0.284(0.84) & 0.325(0.90) &
 0.279(0.86) & 0.321(0.54)\\
 
 & Exchange &
 \textbf{0.352(5.78)} & \textbf{0.398(2.65)} &
 0.477(29.92) & 0.442(11.52) &
 \underline{0.367(22.84)} & \underline{0.407(11.20)} &
 0.442(76.98) & 0.434(27.87) &
 0.422(19.03) & 0.426(7.83) &
 0.402(20.44) & 0.419(6.13)\\
 
 & Weather &
 \textbf{0.245(0.48)} & \underline{0.273(0.31)} &
 \underline{0.246(1.14)} & \textbf{0.272(1.06)} &
 0.258(1.48) & 0.279(1.06) &
 0.253(3.63) & 0.277(2.35) &
 0.264(1.56) & 0.281(0.99) &
 0.247(1.48) & 0.273(1.18)\\\hline
 
\multicolumn{2}{c}{\# Top-2}&
\multicolumn{2}{c}{\textbf{11}} &
\multicolumn{2}{c}{8} &
\multicolumn{2}{c}{3} &
\multicolumn{2}{c}{0} &
\multicolumn{2}{c}{0} &
\multicolumn{2}{c}{2} \\\hline

 \multirow{7}{*}{\rotatebox{90}{FEDformer}}
 & ETTh1 &
 \underline{0.456(2.44)} & \underline{0.473(3.44)} &
 0.594(54.93) & 0.560(35.78) &
 0.559(25.98) & 0.539(11.93) &
 0.642(30.84) & 0.565(22.96) &
 \textbf{0.443(8.40)} & \textbf{0.454(4.72)} &
 0.541(57.84) & 0.514(36.18)\\
 
 & ETTh2 &
 \textbf{0.414(2.15)} & \underline{0.440(2.03)} &
 0.481(26.05) & 0.478(14.44) &
 0.475(19.96) & 0.487(13.37) &
 1.042(262.31) & 0.716(96.97) &
 \underline{0.415(6.17)} & \textbf{0.433(4.29)} &
 0.632(284.16) & 0.537(127.95)\\
 
 & ETTm1 &
 \textbf{0.397(3.16)} & \textbf{0.430(2.55)} &
 0.524(19.81) & 0.505(13.54) &
 0.604(47.20) & 0.539(22.34) &
 0.621(42.69) & 0.548(28.97) &
 \underline{0.439(20.64)} & \underline{0.443(8.25)} &
 0.605(49.18) & 0.551(36.95)\\
 
 & ETTm2 &
 \underline{0.297(1.27)} & \underline{0.350(2.18)} &
 0.324(18.22) & 0.365(10.49) &
 0.354(51.37) & 0.388(30.51) &
 0.757(253.18) & 0.509(73.51) &
 \textbf{0.295(2.56)} & \textbf{0.341(1.46)} &
 1.007(360.70) & 0.671(119.15)\\
 
 & Exchange &
 \textbf{0.466(5.23)} & \textbf{0.473(3.26)} &
 0.546(112.00) & 0.509(45.92) &
 \underline{0.476(82.30)} & \underline{0.490(41.62)} &
 0.609(102.88) & 0.540(38.47) &
 0.524(12.21) & 0.503(5.43) &
 0.588(131.81) & 0.537(70.14)\\
 
 & Weather &
 \underline{0.346(10.33)} & \underline{0.385(8.51)} &
 0.374(26.57) & 0.399(21.56) &
 0.353(28.35) & 0.390(20.00) &
 0.695(120.37) & 0.584(56.48) &
 \textbf{0.306(9.91)} & \textbf{0.351(11.44)} &
 0.521(52.04) & 0.495(24.69)\\\hline
 
 \multicolumn{2}{c}{\# Top-2}&
 \multicolumn{2}{c}{\textbf{12}} &
 \multicolumn{2}{c}{0} &
 \multicolumn{2}{c}{2} &
 \multicolumn{2}{c}{0} &
 \multicolumn{2}{c}{10} &
 \multicolumn{2}{c}{0} \\\hline
 
 \multirow{7}{*}{\rotatebox{90}{PAttn}}
 & ETTh1 &
 \underline{0.453(0.58)} & \underline{0.438(0.33)} &
 0.463(1.64) & 0.442(1.22) &
 0.461(1.58) & 0.444(1.16) &
 0.462(12.08) & 0.442(7.21) &
 \textbf{0.452(5.77)} & \textbf{0.437(4.99)} &
 0.456(6.90) & 0.440(4.31)\\
 
 & ETTh2 &
 \textbf{0.366(0.80)} & \textbf{0.393(0.44)} &
 0.372(8.36) & 0.398(4.78) &
 0.371(2.33) & 0.400(2.22) &
 \underline{0.370(3.03)} & 0.397(2.30) &
 0.372(4.30) &\underline{0.397(2.36)} &
 0.374(4.34) & 0.398(2.68)\\
 
 & ETTm1 &
 \textbf{0.378(0.39)} & \textbf{0.391(0.34)} &
 0.384(3.64) & 0.392(2.58) &
 \underline{0.380(0.90)} & \underline{0.391(0.70)} &
 0.384(1.78) & 0.392(0.89) &
 0.393(6.30) & 0.396(3.66) &
 0.385(2.05) & 0.392(0.80)\\
 
 & ETTm2 &
 \textbf{0.279(0.21)} & \textbf{0.325(0.27)} &
 0.284(2.32) & 0.330(1.90) &
 0.285(0.84) & 0.331(0.84) &
 \underline{0.282(1.13)} & \underline{0.328(0.79)} &
 0.284(1.79) & 0.329(1.63) &
 0.284(1.02) & 0.329(1.08)\\
 
 & Exchange &
 \textbf{0.362(2.20)} & \textbf{0.402(1.46)} &
 0.372(6.31) & 0.411(4.02) &
 0.377(9.72) & 0.411(4.94) &
 0.373(10.13) & 0.410(5.24) &
 0.372(5.66) & 0.410(3.51) &
 \underline{0.370(3.76)} & \underline{0.410(2.48)}\\
 
 & Weather &
 \underline{0.260(0.14)} & \underline{0.279(0.22)} &
 0.259(0.30) & 0.279(0.25) &
 0.260(0.38) & 0.279(0.31) &
 0.260(0.68) & 0.278(0.65) &
 0.261(0.80) & 0.279(0.56) &
 \textbf{0.259(0.41)} & \textbf{0.277(0.28)}\\\hline
 
 \multicolumn{2}{c}{\# Top-2}&
 \multicolumn{2}{c}{\textbf{10}} &
 \multicolumn{2}{c}{1} &
 \multicolumn{2}{c}{2} &
 \multicolumn{2}{c}{4} &
 \multicolumn{2}{c}{3} &
 \multicolumn{2}{c}{4} \\\hline
 
 \multirow{7}{*}{\rotatebox{90}{MultiPatchFormer}}
 & ETTh1 &
 \textbf{0.437(1.88)} & \textbf{0.429(0.97)} &
 0.441(4.54) & 0.433(2.42) &
 0.444(5.03) & 0.437(3.42) &
 0.447(4.17) & 0.438(2.77) &
 \underline{0.441(4.58)} & \underline{0.431(2.89)} &
 0.445(2.92) & 0.432(1.66)\\
 
 & ETTh2 &
 \textbf{0.367(0.78)} & \textbf{0.393(0.65)} &
 \underline{0.369(3.19)} & 0.396(2.58) &
 0.370(3.77) & 0.396(1.92) &
 0.378(3.89) & 0.402(3.08) &
 0.372(3.59) & \underline{0.396(1.82)} &
 0.375(3.32) & 0.398(2.52)\\
 
 & ETTm1 &
 \textbf{0.379(1.15)} & \textbf{0.390(0.69)} &
 0.383(1.32) & 0.392(1.08) &
 \underline{0.381(0.97)} & \underline{0.392(0.89)} &
 0.394(6.36) & 0.400(4.35) &
 0.391(6.17) & 0.396(2.30) &
 0.384(3.14) & 0.392(2.47)\\
 
 & ETTm2 &
 \textbf{0.277(0.87)} & \underline{0.322(0.51)} &
 0.279(2.03) & 0.323(1.66) &
 0.281(1.89) & 0.325(1.24) &
 0.284(2.91) & 0.325(1.75) &
 0.282(2.07) & 0.325(1.11) &
 \underline{0.277(2.26)} & \textbf{0.321(0.98)}\\
 
 & Exchange &
 \textbf{0.360(3.49)} & \textbf{0.404(2.00)} &
 0.378(7.79) & 0.415(4.20) &
 \underline{0.376(7.67)} & \underline{0.412(4.59)} &
 0.439(42.88) & 0.436(13.52) &
 0.377(14.65) & 0.414(7.94) &
 0.378(5.20) & 0.414(3.23)\\
 
 & Weather &
 \textbf{0.243(1.05)} & \textbf{0.268(0.56)} &
 0.254(1.33) & 0.274(1.17) &
 0.254(0.80) & 0.274(0.73) &
 0.246(0.80) & 0.271(0.48) &
 0.258(2.24) & 0.277(1.29) &
 \underline{0.245(1.27)} & \underline{0.270(1.51)}\\\hline
 
 \multicolumn{2}{c}{\# Top-2}&
 \multicolumn{2}{c}{\textbf{12}} &
 \multicolumn{2}{c}{1} &
 \multicolumn{2}{c}{4} &
 \multicolumn{2}{c}{0} &
 \multicolumn{2}{c}{3} &
 \multicolumn{2}{c}{4} \\\hline

\multirow{3}{*}{\rotatebox{90}{t-PG}}
& Activity$^\ddag$ &
\textbf{0.00267(0.01)} &\textbf{ 0.00317(0.09)} &
0.00275(0.13) & 0.00321(0.37) &
0.00272(0.11) & 0.00322(1.04) &
0.00285(0.25) & 0.00331(2.54) &
0.00679(6.93) & 0.00532(35.40) &
\underline{0.00272(0.02)} & \underline{0.00321(0.32)}\\

& PhysioNet$^\ddag$ &
\textbf{0.00496(0.03)} & \underline{0.00381(1.11)} &
0.00512(0.12) & 0.00382(1.65) &
0.00514(0.18) & \textbf{0.00377(0.96)} &
0.00637(2.29) & 0.00430(9.74) &
0.00549(0.17) & 0.00394(1.52) &
\underline{0.00509(0.09)} & 0.00383(0.96)\\
& USHCN$^\ddag$ &
\textbf{0.495(2.00)} & \underline{0.305(5.00)} &
\underline{0.497(4.00)} & 0.309(3.00) &
0.502(2.00) & 0.312(10.0) &
0.499(2.00) & 0.313(8.00) &
0.507(3.00) & 0.315(3.00) &
0.499(3.00) & \textbf{0.305(10.0)}\\\hline

\multicolumn{2}{c}{\# Top-2}&
\multicolumn{2}{c}{\textbf{6}} &
\multicolumn{2}{c}{1} &
\multicolumn{2}{c}{1} &
\multicolumn{2}{c}{0} &
\multicolumn{2}{c}{0} &
\multicolumn{2}{c}{4} \\\hline
 
\multicolumn{2}{c}{Total \# Top-2}&
\multicolumn{2}{c}{\textbf{84}} &
\multicolumn{2}{c}{13} &
\multicolumn{2}{c}{24} &
\multicolumn{2}{c}{15} &
\multicolumn{2}{c}{20} &
\multicolumn{2}{c}{24} \\\hline
\end{tabular}}
\label{table_main}
\end{table*}

\begin{table}[!t]
\caption{
Complexity and efficiency of attention modules with the TimeXer backbone under fixed dimensionality. 
The patching technique \cite{nie_time_2023} is used with patch size 16.
}
\centering
\resizebox{1\columnwidth}{!}{
\begin{tabular}{cccrrrrr}
\hline
\multirow{2}{*}{Attention Method} & Computational & Memory & FLOPs & \# Params \\
& Complexity & Complexity & (M) & (K)\\
\hline
SA & $O(n^2)$ & $O(n^2)$ & 2.398 & 197.376\\ 
Geometry Attention & $O(n^2)$ & $O(n^2)$ & 2.407 & 197.376\\
ProbSparse Attention & $O(n\cdot\text{log} n)$ & $O(n\cdot\text{log} n)$ & 2.414 & 197.376\\
AutoCorrelation & $O(n\cdot\text{log} n)$ & $O(n\cdot\text{log} n)$ & 2.400 & 197.376\\
TSSA & $O(n)$ & $O(1)$ & 1.613 & 139.826\\
SGA* & $O(n)$ & $O(n)$ & \textbf{0.820} & \textbf{67.300}\\
\hline
\end{tabular}}
\label{table_efficiency}
\end{table}

Forecasting results are summarized in Table~\ref{table_main}. 
For each backbone, we follow the hyper-parameter configuration in its official implementation whenever available. 
No extra hyper-parameter tuning is performed for SGA. 
TimeXer, iTransformer, CARD, FEDformer, PAttn, MultiPatchFormer, and t-PatchGNN (t-PG) use SA as the default attention module, while SimpleTM uses Geometry attention.

\textbf{Main Results on Regular Time Series:}
Table~\ref{table_main} shows that SGA obtains the largest number of best or second-best results on each regular time series backbone. 
It achieves 12, 9, 12, 11, 12, 10, and 12 best or second-best results on TimeXer, iTransformer, SimpleTM, CARD, FEDformer, PAttn, and MultiPatchFormer, respectively. 
Compared with the default attention module of each backbone, SGA reduces MSE by up to about 26\%.

\textbf{Main Results on Irregular Time Series:}
As shown at the bottom of Table~\ref{table_main}, SGA achieves the lowest MSE on the irregular clinical, human activity, and climate datasets. 
On the official t-PatchGNN backbone, SGA achieves about 8\% lower average MSE than the other attention methods.

\begin{figure}[!t]
\centering
\includegraphics[width=1.0\columnwidth]{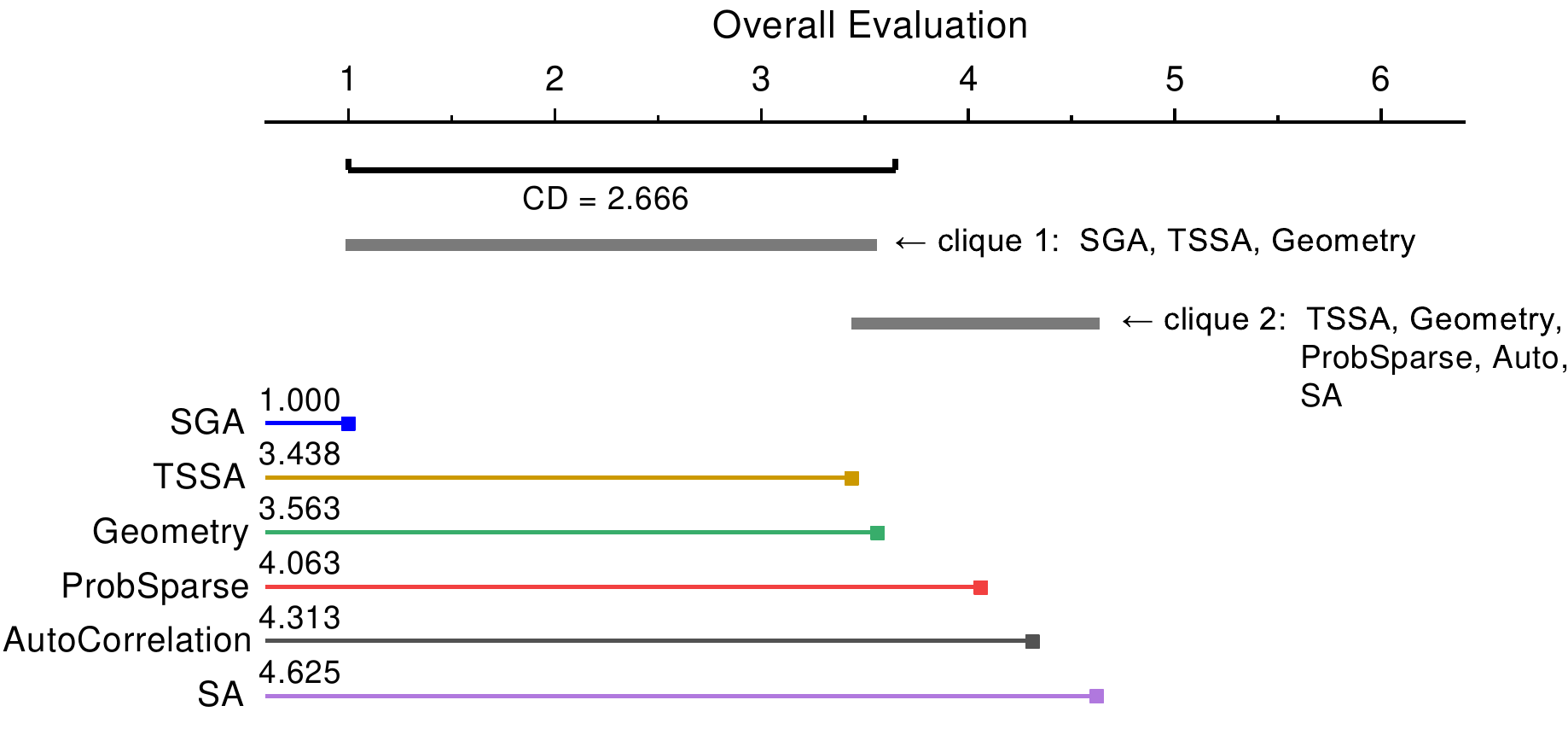}
\caption{Nemenyi critical difference diagram. Horizontal bars connect methods whose differences are not statistically significant.}
\label{nemenyiCD}
\end{figure}

Overall, SGA achieves 84 best or second-best results out of 90 cases, ranking first among all attention methods. 
Geometry and TSSA rank second with 24 top-two results each, while SA obtains 13 top-two results. 
To further compare different attention methods, we report the average rank and the Nemenyi critical difference diagram in Fig.~\ref{nemenyiCD}. 

In this analysis, each backbone is treated as one statistical unit. 
For each backbone, attention methods are ranked according to the number of best or second-best results, and tied methods are assigned the average rank. 
Let $\mathcal{B}$ denote the set of evaluated backbones and let $B=|\mathcal{B}|$.
Let $r_{m,b}$ denote the rank of attention method $m$ on backbone $b$. 
The average rank of method $m$ is calculated as:
\begin{equation}
\mathrm{AvgRank}(m)=\frac{1}{B}\sum_{b\in\mathcal{B}} r_{m,b}.
\end{equation}
A lower average rank indicates better overall performance.

Based on the updated results over eight backbones, SGA obtains the lowest average rank of 1.000, followed by TSSA (3.438), Geometry (3.563), ProbSparse (4.063), AutoCorrelation (4.313), and SA (4.625). 
The critical difference in the Nemenyi test is computed as:
\begin{equation}
\mathrm{CD}=q_{\alpha}\sqrt{\frac{M(M+1)}{6B}},
\end{equation}
where $M$ is the number of compared attention methods, $B$ is the number of evaluated backbones, and $q_{\alpha}$ is the critical value at significance level $\alpha$. 
In our setting, $M=6$, $B=8$, and $q_{0.05}=2.850$, giving $\mathrm{CD}=2.666$. 
Therefore, SGA is significantly better than ProbSparse, AutoCorrelation, and SA. These results show that SGA achieves strong overall performance across different backbones.

Together with the forecasting results in Table \ref{table_main}, these efficiency results indicate that SGA maintains competitive performance while using fewer parameters and FLOPs.
These results suggest that simplified attention computation can still maintain competitive forecasting performance.

\begin{table}[!t]
\caption{
Ablation study based on the TimeXer backbone. 
Mean MSE and MAE over all prediction lengths are reported.}
\centering
\resizebox{1\columnwidth}{!}{
\begin{tabular}{ccccccccccccc}
\hline
\multirow{2}{*}{Dataset} & 
\multicolumn{2}{c}{ETTh1} & 
\multicolumn{2}{c}{ETTh2} & 
\multicolumn{2}{c}{ETTm1} & 
\multicolumn{2}{c}{ETTm2} & 
\multicolumn{2}{c}{Exchange} & 
\multicolumn{2}{c}{Weather} \\
& 
MSE & MAE & 
MSE & MAE & 
MSE & MAE & 
MSE & MAE & 
MSE & MAE & 
MSE & MAE \\\hline

$\textit{w/o}$ $\bm{\mathcal{R}}_t$ & 
0.448 & 0.435 & 
0.378 & 0.398 & 
0.381 & 0.390 & 
0.276 & 0.320 & 
0.383 & 0.413 & 
0.242 & 0.268\\

$\textit{w/o}$ $\bm{\mathcal{A}}$ & 
0.445 & 0.436 & 
0.368 & 0.395 & 
0.378 & 0.389 & 
0.274 & 0.318 & 
0.373 & 0.410 & 
0.238 & 0.265\\

$\textit{w/o}$ Sparse & 
0.442 & 0.436 & 
0.369 & 0.396 & 
0.383 & 0.393 & 
0.271 & 0.317 & 
0.380 & 0.414 & 
0.239 & 0.265\\

$\textit{w/o}$ Orth. & 
0.443 & 0.436 & 
0.369 & 0.394 & 
0.380 & 0.390 & 
0.272 & 0.317 & 
0.382 & 0.413 & 
0.239 & 0.266\\

$\textit{re}$ MLP & 
0.448 & 0.433 & 
0.369 & 0.394 & 
0.381 & 0.389 & 
0.272 & 0.317 & 
0.383 & 0.413 & 
0.241 & 0.266\\

$\textit{re}$ SA & 
0.452 & 0.436 & 
0.369 & 0.395 & 
0.381 & 0.390 & 
0.273 & 0.318 & 
0.381 & 0.413 & 
0.240 & 0.266\\

SGA & 
\textbf{0.435} & \textbf{0.431} & 
\textbf{0.366} & \textbf{0.393} & 
\textbf{0.378} & \textbf{0.388} & 
\textbf{0.271} & \textbf{0.317} & 
\textbf{0.363} & \textbf{0.404} & 
\textbf{0.237} & \textbf{0.264}\\\hline
\end{tabular}}

\label{ablation}
\end{table}

\subsection{Ablation Study}
\label{ablationstudy}
To validate the design of SGA, we conduct ablation experiments on the TimeXer backbone across all six regular datasets. 
No additional hyper-parameter tuning is performed, and the configuration follows the TimeXer backbone for a consistent comparison. 
We compare the complete SGA with several variants. 
\textit{w/o} $\bm{\mathcal{R}}_t$, \textit{w/o} $\bm{\mathcal{A}}$, \textit{w/o} Sparse, and \textit{w/o} Orth. remove the input-dependent residual score matrix, the shared attention score matrix, Top-$K$ sparsification, and orthogonal initialization, respectively. 
\textit{re} MLP and \textit{re} SA replace the residual score matrix with MLP-based and self-attention-based constructions. 
The results are summarized in Table~\ref{ablation}.

Removing $\bm{\mathcal{R}}_t$ leads to worse performance on all datasets, showing that the input-dependent residual score matrix is important for capturing local variations. 
Removing $\bm{\mathcal{A}}$ also degrades performance in most cases, which indicates the benefit of the shared attention score matrix. 
Removing Top-$K$ sparsification leads to worse results in most settings, suggesting that sparsification helps reduce weakly related or redundant positions. The variants \textit{re} MLP and \textit{re} SA do not achieve the best result in any setting. 
This suggests that the proposed normalized-energy-based residual construction provides a more suitable lightweight cue for SGA than the tested MLP-based and self-attention-based residual constructions. The result of \textit{w/o} Orth. shows that orthogonal initialization helps different heads start from more diverse shared score matrices. 
Overall, these results show that the main components generally contribute to the final performance of SGA.

\subsection{Computational and Memory Efficiency}
We conduct an efficiency comparison to evaluate the computational advantage of SGA in resource-constrained forecasting settings. 
With the patch technique, we set the look-back length to $n$ and the prediction length to $s$. 
The model dimension is fixed at $d=512$, and the batch size is fixed at 32.

From Eq.~\eqref{eq_sga_def}, SGA first projects the input into $\bm{V}_t$ and then aggregates $\bm{V}_t$ using the fused attention score matrix $\psi(\bm{\mathcal{A}}, \bm{\mathcal{R}}_t)$. 
Since SGA avoids query and key projections, it reduces both parameter count and FLOPs compared with standard SA. 

Table~\ref{table_efficiency} reports the complexity comparison among different attention methods. The complexity in Table \ref{table_efficiency} is reported in terms of the look-back length $n$, with the prediction length $s$ fixed. The complexity values of the compared methods are taken from their original papers. 
As shown in the table, SGA uses fewer FLOPs and fewer parameters than standard SA and the compared efficient attention variants. 
This is mainly because SGA keeps the value projection but removes the query and key projections.

\begin{figure}
\centering
\includegraphics[width=0.45\columnwidth]{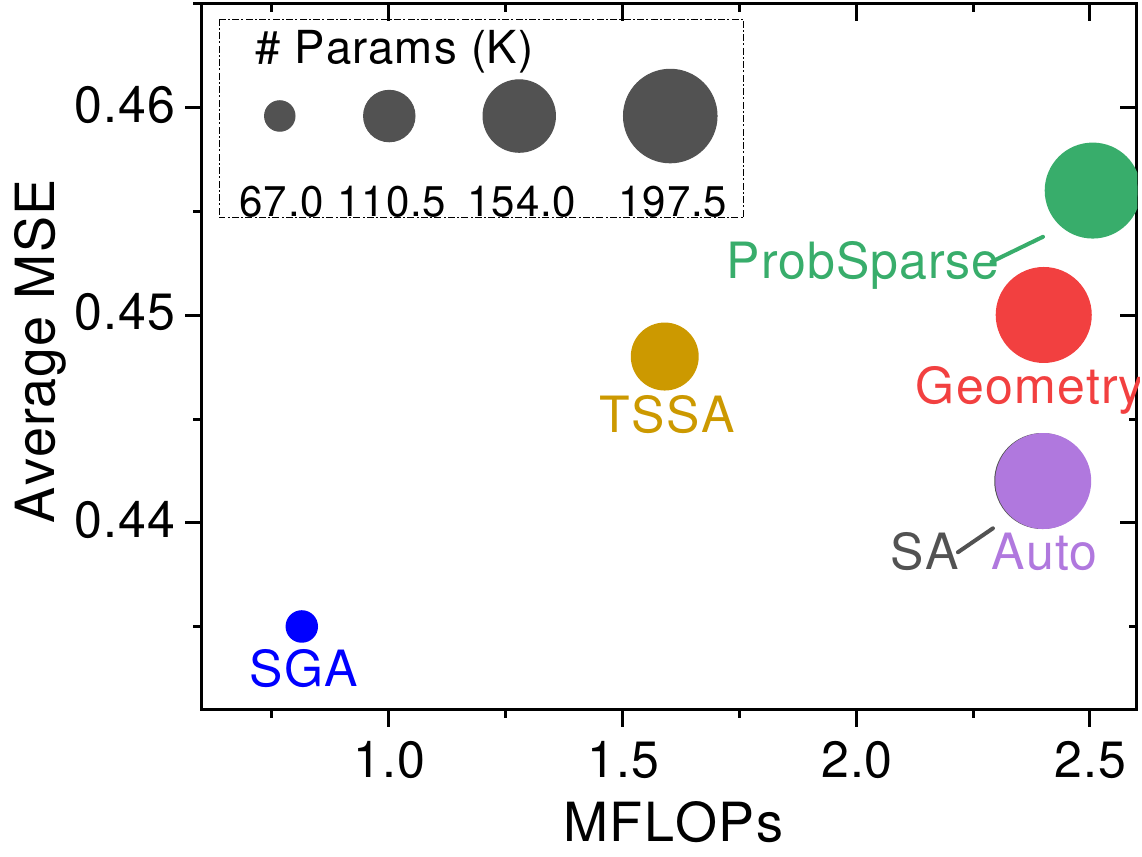}
\includegraphics[width=0.45\columnwidth]{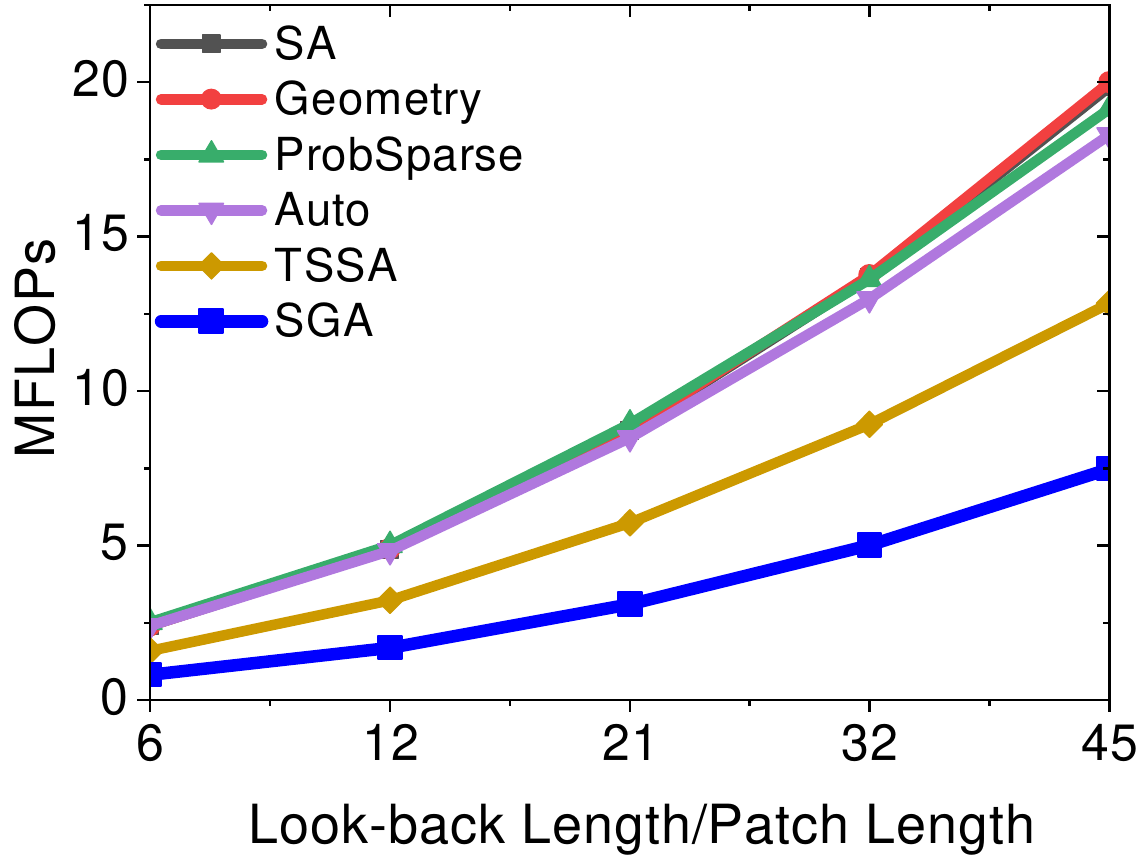}
\includegraphics[width=0.45\columnwidth]{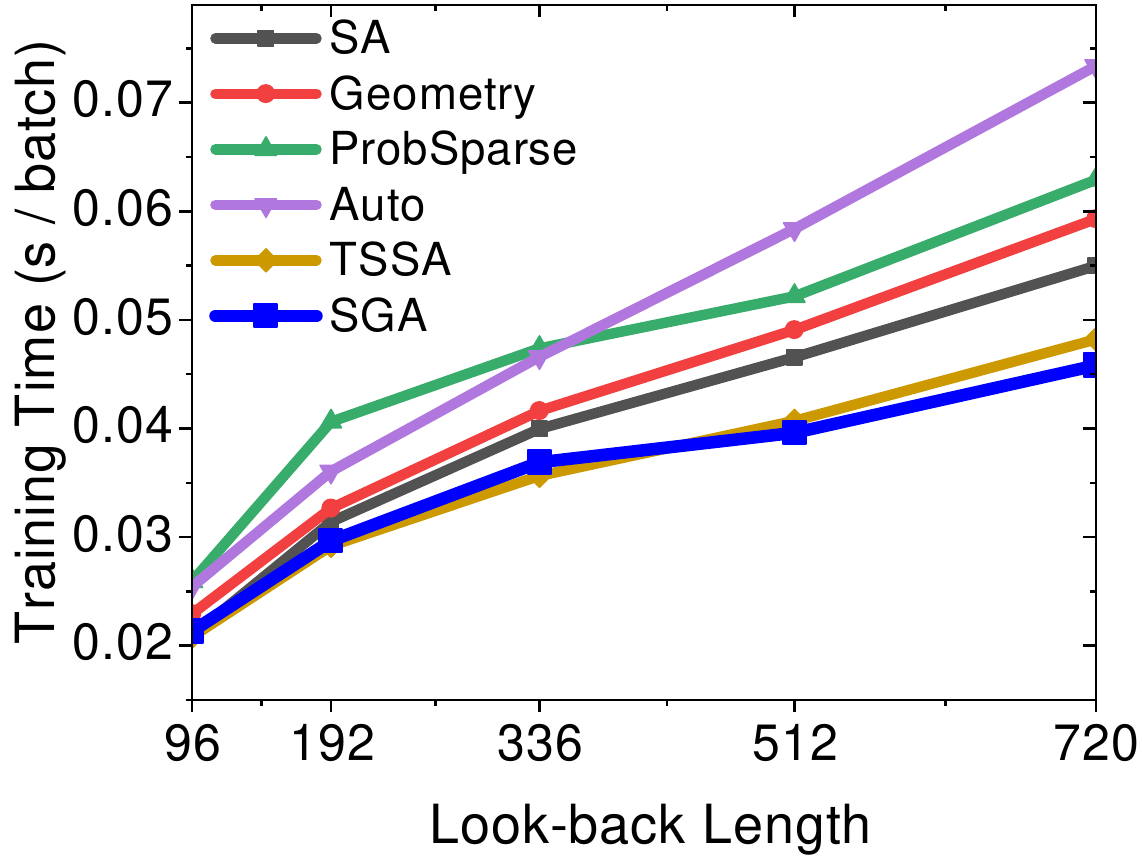}
\includegraphics[width=0.45\columnwidth]{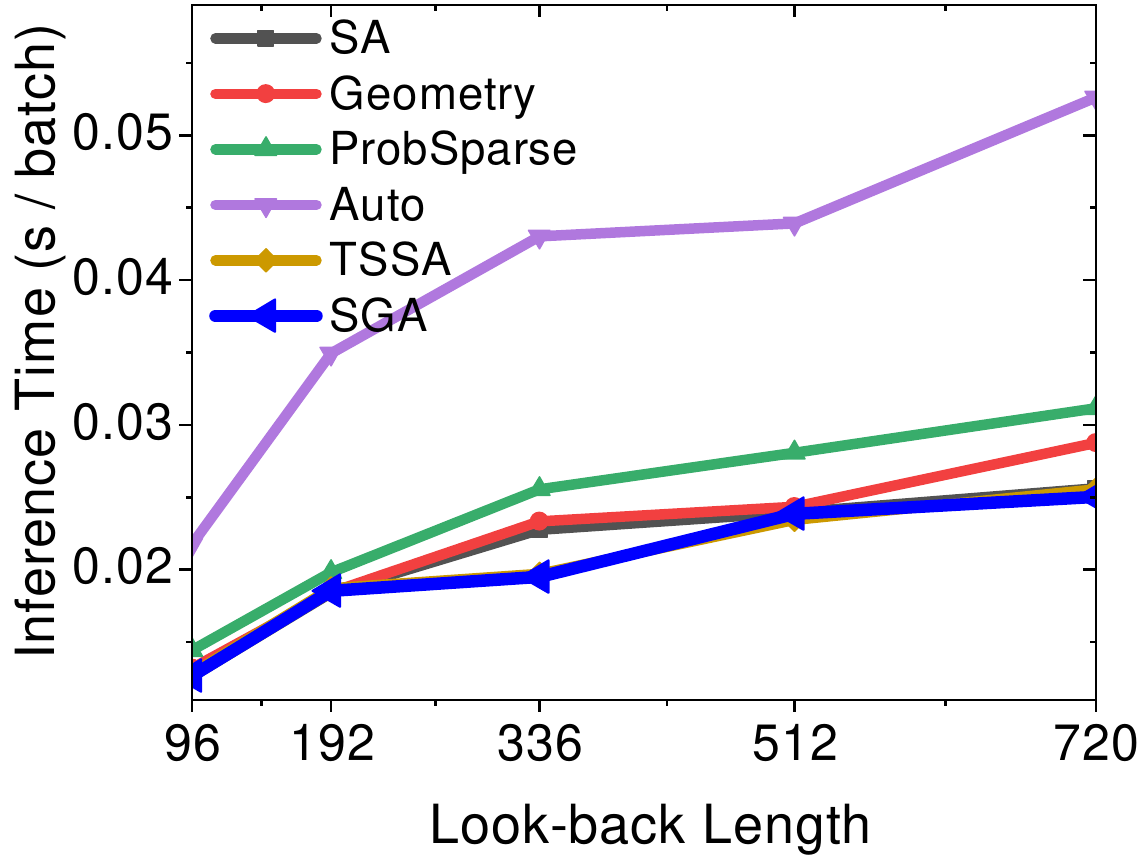}
\caption{
Efficiency comparison with the TimeXer backbone on the ETTh1.}
\label{efficiency}
\end{figure}

Fig.~\ref{efficiency} further compares FLOPs, training time, and inference time with the TimeXer backbone on the ETTh1. 
The upper-left panel summarizes the overall efficiency comparison. 
The upper-right panel reports FLOPs under different look-back lengths. 
Under the patch technique, the dominant cost comes from linear projections. 
Since SGA does not compute query or key projections, its FLOPs grow more slowly than standard SA as the look-back length increases.

The bottom panels report training time and inference time in seconds per batch. 
SGA achieves near-best training and inference efficiency across different look-back lengths. 
On average, SGA trains 1.19x faster and runs inference 1.25x faster than the compared attention methods. 
These results support the efficiency advantage of SGA for resource-constrained forecasting settings.

\subsection{Increasing the Look-back Length}

Here, we examine the effect of using longer historical input sequences. 
To keep the comparison consistent, all experiments in this section follow a unified setting based on \cite{wang_timexer_2024, zhou_informer_2021}. 
The number of encoder layers is set to 2, the model dimension is 512, the batch size is 32, and the learning rate is 0.0001. 
No additional hyper-parameter tuning is performed.

\begin{figure}[t!]
\centering
\includegraphics[width=.92\columnwidth]{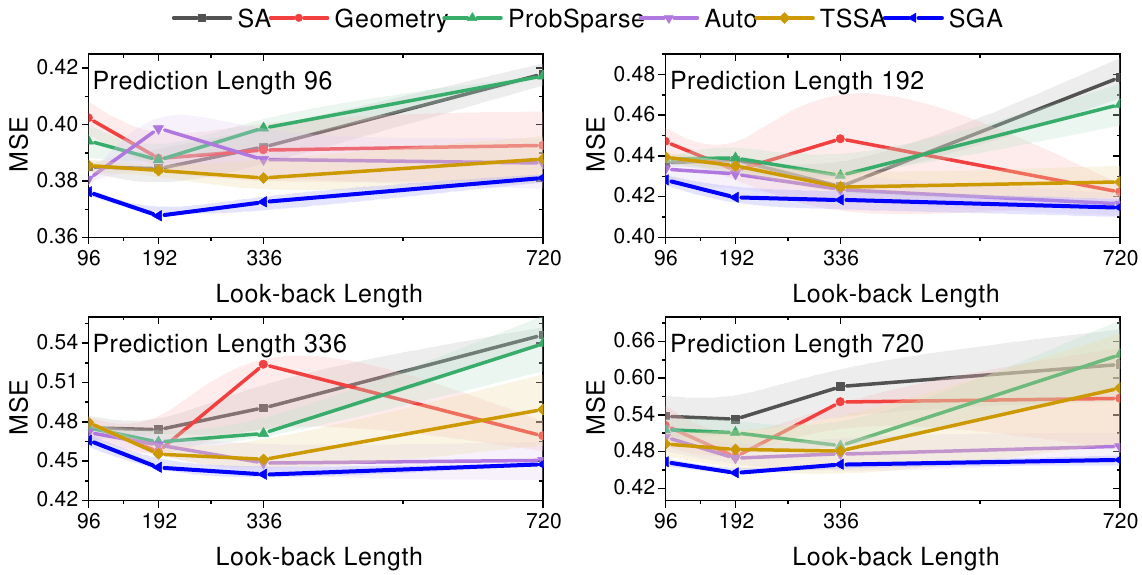}
\caption{Performance comparison on the ETTh1 with the TimeXer backbone under different look-back lengths $n \in \{96, 192, 336, 512, 720\}$. The shading denotes the standard deviation of MSE fitted with a B-spline.}
\label{seq_len}
\end{figure}

As shown in Fig.~\ref{seq_len}, SGA maintains competitive performance as the look-back length increases, and shows stronger performance than the compared attention methods under longer input settings. 
For most methods, the performance first improves and then declines as the look-back length increases. 
A longer input window can provide more useful historical information at the beginning. 
However, when the look-back length becomes too large, extra noise may be introduced, which can reduce forecasting performance. 
This trend is also accompanied by a larger standard deviation, indicating higher performance uncertainty.

\begin{figure}[t]
\centering
\includegraphics[width=0.45\columnwidth]{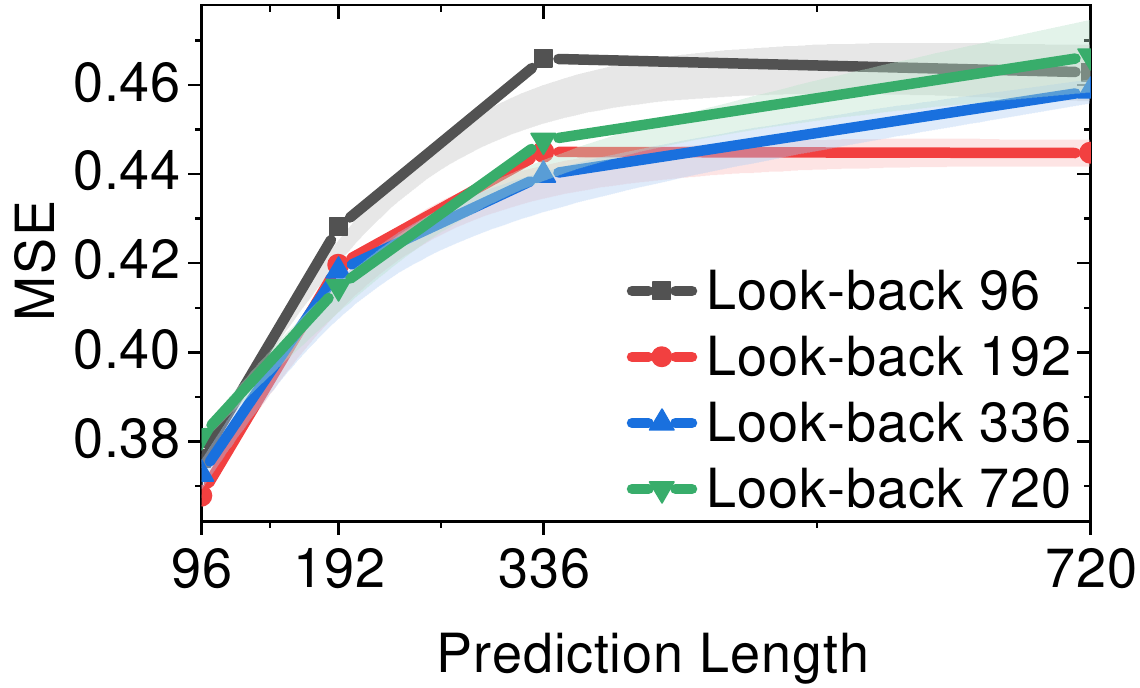}
\llap{(a)~}
\includegraphics[width=0.45\columnwidth]{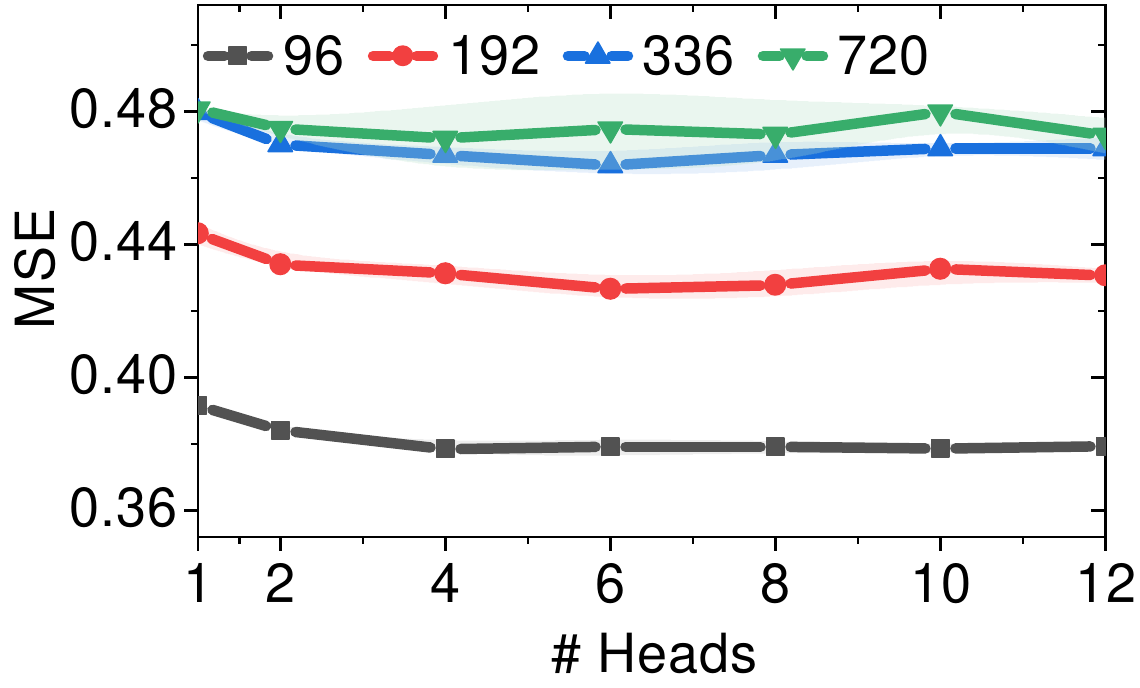}
\llap{(b)~}
\caption{(a) Performance of SGA with different look-back lengths on the ETTh1 using the TimeXer backbone. The shaded region represents the standard deviation of MSE fitted by a B-spline. (b) Performance of SGA with different numbers of attention heads on the ETTh1.}
\label{fig_inputlen_nhead}
\end{figure}

Fig.~\ref{fig_inputlen_nhead} (a) further shows that compared with a look-back length of 96, a length of 192 gives better performance. 
When the look-back length is further increased, the gain becomes limited, and excessively long input windows may even reduce performance because of additional noise.

\subsection{Effect of the Number of Attention Heads}

We study the effect of the number of attention heads in Fig.~\ref{fig_inputlen_nhead} (b). 
The experiments are conducted with the TimeXer backbone on the ETTh1. 
The number of attention heads varies from 1 to 12. To keep the value dimension of each head fixed, the model dimension changes with the number of heads.
The dimension of the feed-forward network is set to four times the value dimension.

As shown in Fig.~\ref{fig_inputlen_nhead} (b), increasing the number of attention heads improves the performance of SGA at first. 
However, when the number of heads exceeds 4, the improvement becomes limited. 
This result suggests that a small number of attention heads is sufficient for SGA in this setting, and adding more heads may introduce redundant computation.

\subsection{Effect of Hyper-parameters}

\begin{figure}[t!]
\centering
\includegraphics[width=\columnwidth]{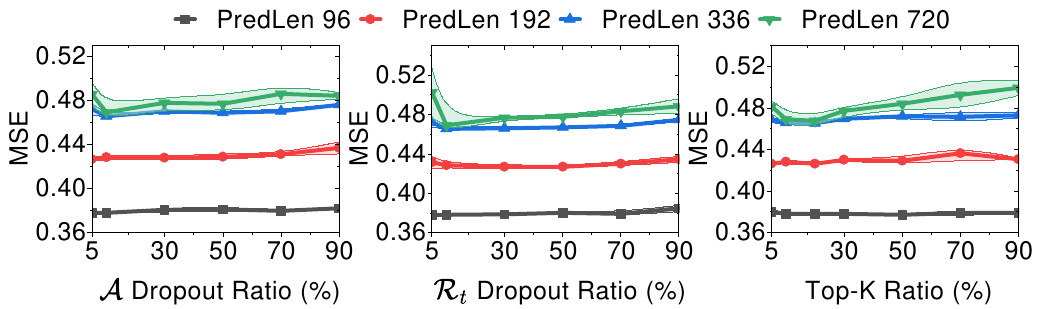}
\caption{
Sensitivity analysis of SGA hyper-parameters on the ETTm1 using the TimeXer backbone.
}
\label{hyper-parameters}
\end{figure}

SGA has three main hyper-parameters: the Top-$K$ ratio, the dropout ratio for the shared attention score matrix $\bm{\mathcal{A}}$, and the dropout ratio for the input-dependent residual score matrix $\bm{\mathcal{R}}_t$. 
We use the TimeXer backbone on the ETTm1 for this study. 
The results are reported under four prediction lengths, including 96, 192, 336, and 720. 
As shown in Fig.~\ref{hyper-parameters}, SGA shows stable performance under different hyper-parameter settings.

The results show that SGA is generally stable across different Top-$K$ ratios and dropout ratios. 
For prediction length 96, changing the Top-$K$ ratio has little effect on MSE. 
For longer prediction lengths, a very large Top-$K$ ratio may increase MSE because more weakly related positions are retained. This result further supports the redundancy observation in attention scores. 
Keeping too many positions may retain weakly related information, while a moderate Top-$K$ ratio is sufficient for effective attention aggregation.
The shared matrix $\bm{\mathcal{A}}$ is also robust to dropout, although a large dropout ratio may weaken common attention patterns. 
For $\bm{\mathcal{R}}_t$, a moderate dropout ratio helps regularize input-dependent variations, while excessive dropout may remove useful residual information.

\subsection{Visualization of Prediction Results}
\label{visualization}
\begin{figure}[t]
\centering
\includegraphics[width=0.45\columnwidth]{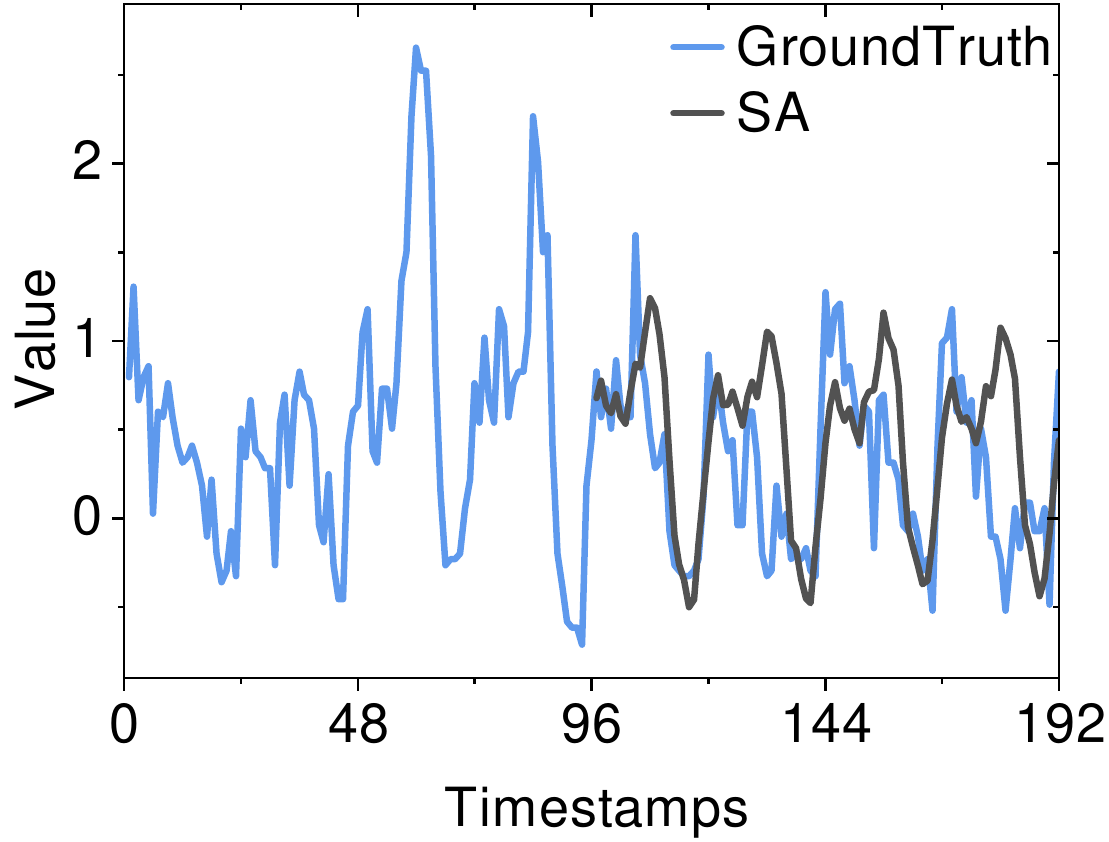}
\llap{(a)~}
\includegraphics[width=0.45\columnwidth]{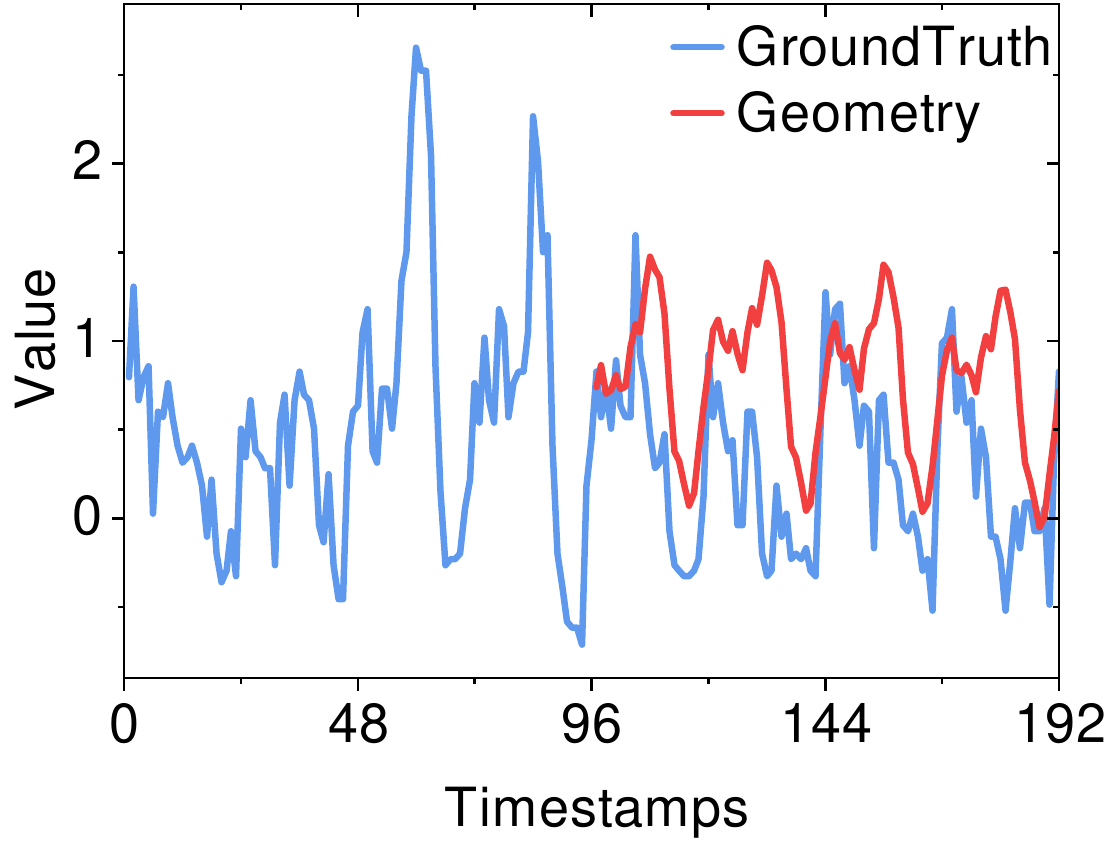}
\llap{(b)~}
\includegraphics[width=0.45\columnwidth]{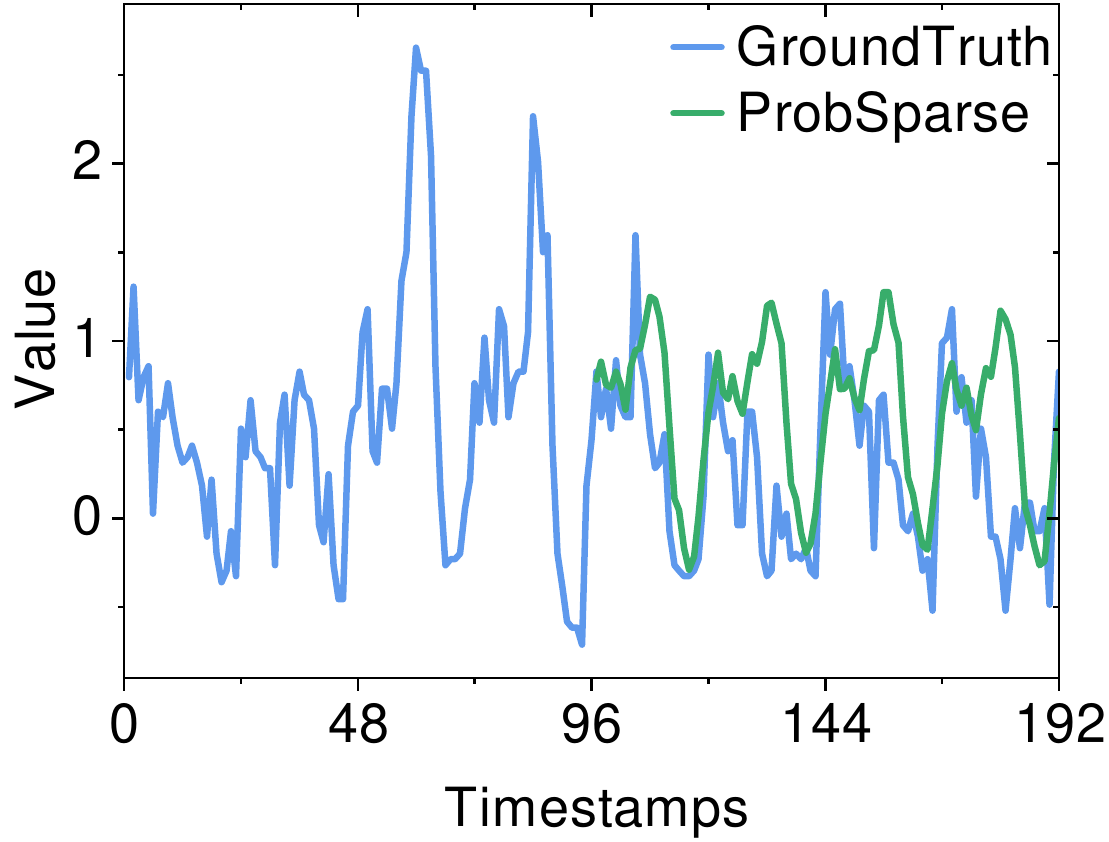}
\llap{(c)~}
\includegraphics[width=0.45\columnwidth]{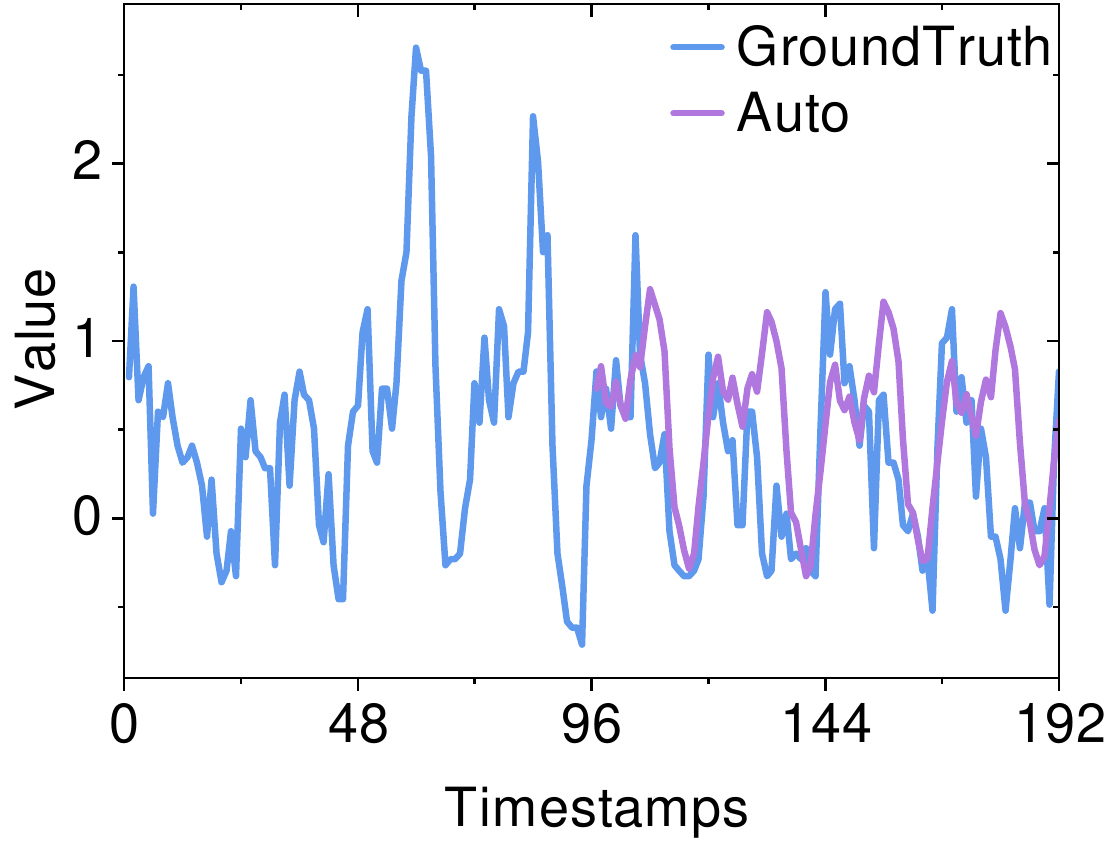}
\llap{(d)~}
\includegraphics[width=0.45\columnwidth]{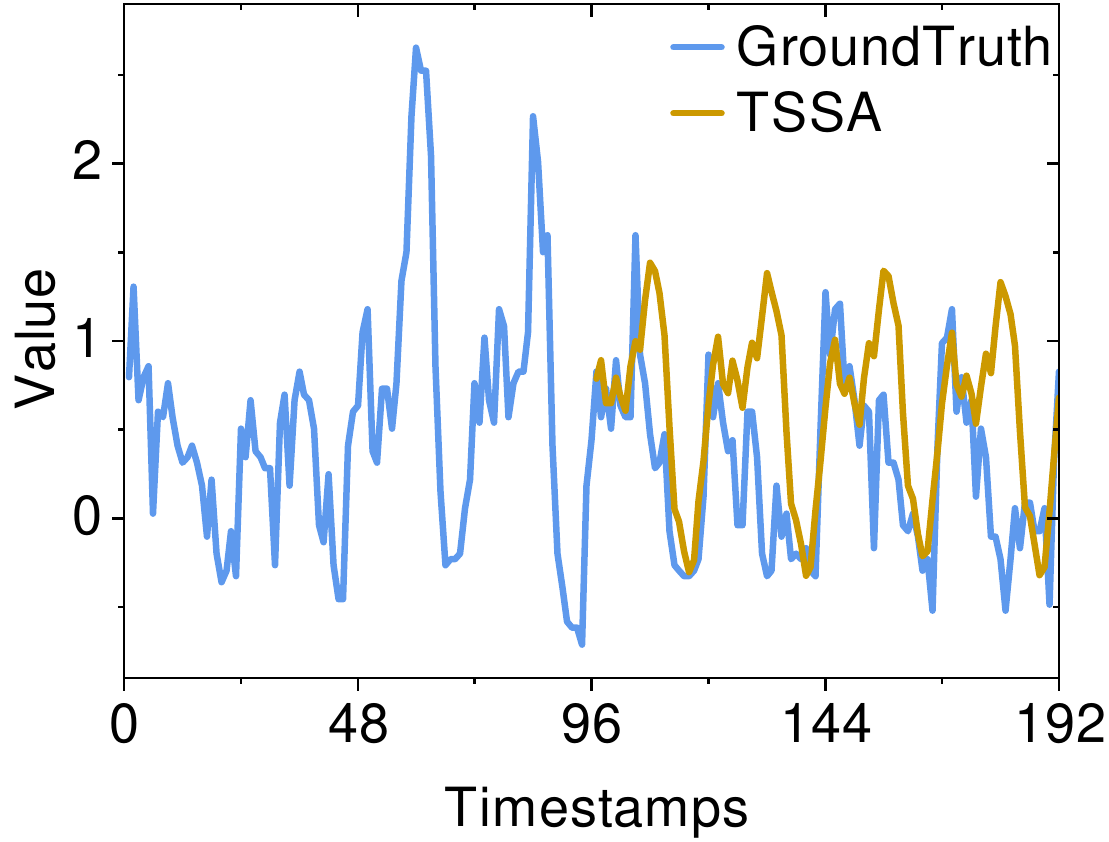}
\llap{(e)~}
\includegraphics[width=0.45\columnwidth]{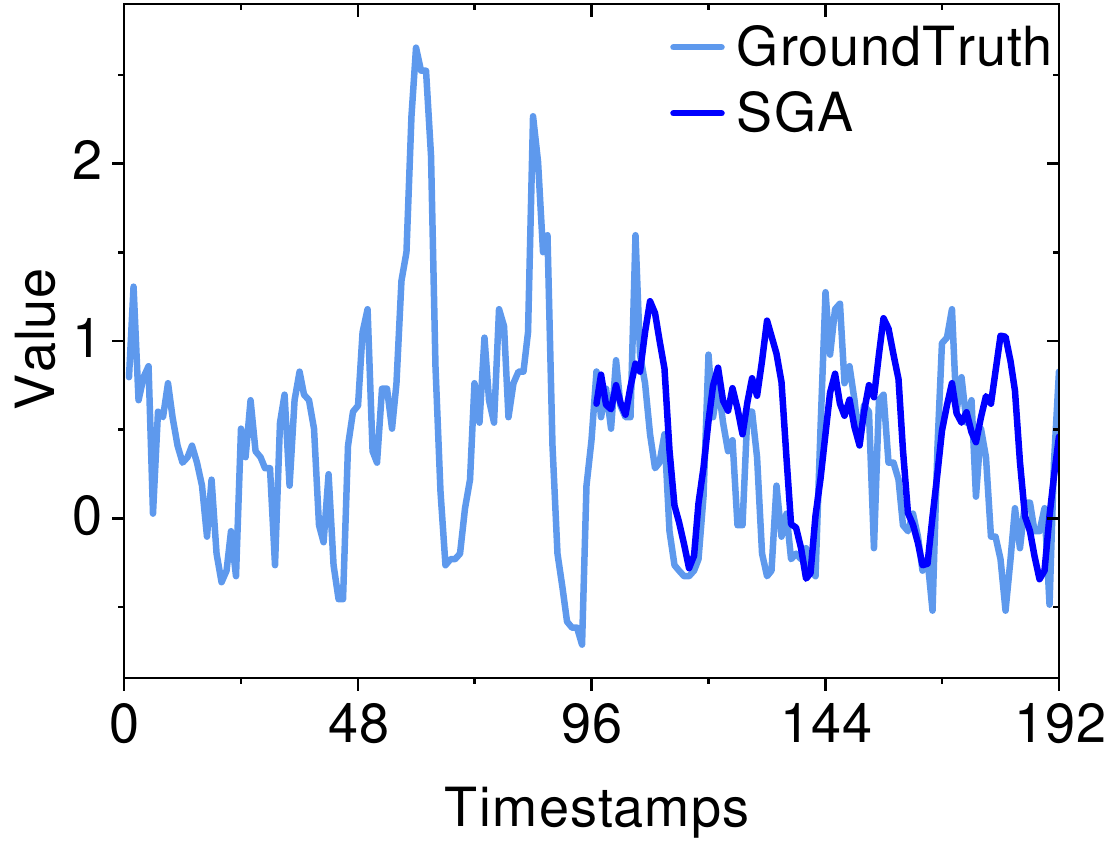}
\llap{(f)~}
\caption{Prediction visualization of six attention mechanisms with the same TimeXer backbone on the ETTh1.}
\label{fig_visualization}
\end{figure}

Fig.~\ref{fig_visualization} compares six attention mechanisms with the same TimeXer backbone on ETTh1. In this case, SGA tracks the main fluctuations more closely than the other mechanisms, while Geometry produces a smoother curve and deviates more from short-term changes. All variants use the original TimeXer training script, whose hyper-parameters are mainly set for SA and are kept unchanged for other mechanisms without separate tuning. Under this setting, SGA obtains the lowest MAE and MSE, with values of 0.368 and 0.250. Compared with SA, SGA reduces MAE and MSE by 0.099\% and 0.545\%, while using only about one third of its parameters. By contrast, Auto, ProbSparse, TSSA, and Geometry perform worse than the TimeXer baseline, with performance drops ranging from -4.365\% to -47.812\% in MAE and from -11.820\% to -89.239\% in MSE. These results suggest that SGA offers a better balance between prediction accuracy and model compactness under this shared training setting.

\section{Conclusion}\label{conclusion}

We observe that SA score maps contain shared patterns across timestamps as well as input-dependent variations. This indicates redundancy in repeated query-key score computation. Motivated by this observation, we propose SGA, which replaces query-key score computation with a shared score matrix and an input-dependent residual score matrix. This design captures common patterns and local variations while achieving linear time and score-matrix memory complexity with respect to the look-back length. Experiments on regular and irregular time series show that SGA maintains competitive forecasting accuracy while improving attention efficiency across different backbones and forecasting horizons. Across all evaluated settings, SGA obtains 84 best or second-best results out of 90 cases. With the TimeXer backbone, SGA reduces FLOPs by over 60\% and parameters by over 60\% compared with SA. These results suggest that SGA provides a favorable balance between accuracy and efficiency and can serve as a plug-and-play attention module for different forecasting backbones. Since the current experiments are conducted on public benchmarks and a high-end GPU platform, the results should be interpreted as benchmark-based efficiency evidence. Extending SGA to complete industrial forecasting systems and broader real-world settings is left for our future work.

\section*{Acknowledgments}
This work is supported by the National Natural Science Foundation of China (No. 12433011), the Guangdong Basic and Applied Basic Research Foundation (No. 2024A1515011962), the Australian Research Council (Nos. FT210100624, DP260100326, DE230101033, DP240101814, LP230200892, and LP240200546).

\appendices

\renewcommand{\thefigure}{\thesection\arabic{figure}}  
\renewcommand{\thetable}{\thesection\arabic{table}}  
\setcounter{figure}{0}  
\setcounter{table}{0}

\section{Detailed Horizon-wise Forecasting Results}
\label{app_detailed_results}
This appendix reports the horizon-wise forecasting results. In the main text, the results are averaged across four prediction lengths for compact presentation, whereas the appendix provides detailed results for assessing the stability of each attention mechanism across different prediction ranges. The horizon-wise results reported here are obtained using the training scripts of the corresponding baseline backbones.

Seven representative backbones are included: TimeXer, iTransformer, SimpleTM, CARD, FEDformer, PAttn, and MultiPatchFormer. For each backbone, the \# Top-2 row reports the total number of best and second-best results across all dataset, prediction-length, and metric combinations. These horizon-wise results show that SGA obtains the highest \# Top-2 count on six of the seven backbones, including TimeXer, iTransformer, SimpleTM, CARD, PAttn, and MultiPatchFormer. On FEDformer, AutoCorrelation obtains the highest count, while SGA ranks second. Across all seven backbones, SGA obtains 284 Top-2 results in total, which is higher than SA, Geometry, ProbSparse, AutoCorrelation, and TSSA. These results suggest that SGA tends to maintain stable forecasting performance across different prediction lengths and backbone models. The corresponding results are reported in Tables~\ref{complete_results_timexer}-\ref{complete_results_multipatchformer}. The best and second-best values are shown in bold and underlined, respectively, based on the unrounded results. `D' denotes the dataset, `PL' denotes the prediction length, and `\# Top-2' denotes the total number of best and second-best results.

\begin{table}[h]
\caption{Complete results based on \textit{TimeXer} backbone. }
\centering
\resizebox{1\columnwidth}{!}{
\begin{tabular}{ccccccccccccccc}
\hline
\multirow{30}{*}{\rotatebox{90}{TimeXer}}&\multirow{2}{*}{D} & \multirow{2}{*}{PL} & \multicolumn{2}{c}{SGA} & \multicolumn{2}{c}{SA} & \multicolumn{2}{c}{Geometry} & \multicolumn{2}{c}{ProbSparse} & \multicolumn{2}{c}{AutoCorrelation} & \multicolumn{2}{c}{TSSA}\\
&&& MSE & MAE & MSE & MAE & MSE & MAE & MSE & MAE & MSE & MAE & MSE & MAE\\\cline{2-15}
& \multirow{4}{*}{\rotatebox{90}{ETTh1}} & 96 & \textbf{0.378} & \underline{0.392} & 0.385 & 0.398 & 0.386 & 0.398 & 0.390 & 0.402 & \underline{0.380} & \textbf{0.391} & 0.386 & 0.395\\
& & 192 & \textbf{0.428} & 0.426 & \underline{0.430} & 0.429 & 0.437 & 0.433 & 0.435 & 0.429 & 0.436 & \textbf{0.425} & 0.438 & \underline{0.426}\\
& & 336 & \textbf{0.466} & \underline{0.443} & \underline{0.471} & 0.448 & 0.477 & 0.454 & 0.479 & 0.450 & 0.475 & \textbf{0.441} & 0.484 & 0.447\\
& & 720 & \textbf{0.469} & \textbf{0.464} & 0.482 & 0.472 & 0.498 & 0.482 & 0.518 & 0.493 & \underline{0.479} & \underline{0.465} & 0.484 & 0.469\\\cline{2-15}

& \multirow{4}{*}{\rotatebox{90}{ETTh2}} & 96 & \textbf{0.280} & \textbf{0.333} & \underline{0.282} & \underline{0.333} & 0.284 & 0.336 & 0.286 & 0.338 & 0.292 & 0.342 & 0.292 & 0.340\\
& & 192 & \textbf{0.360} & \textbf{0.384} & 0.363 & 0.387 & 0.361 & 0.387 & \underline{0.361} & 0.387 & 0.362 & \underline{0.386} & 0.371 & 0.389\\
& & 336 & \textbf{0.411} & \textbf{0.422} & 0.417 & 0.427 & 0.424 & 0.428 & 0.422 & 0.427 & \underline{0.415} & \underline{0.424} & 0.441 & 0.434\\
& & 720 & \textbf{0.412} & \underline{0.434} & 0.433 & 0.445 & 0.433 & 0.447 & \underline{0.413} & \textbf{0.432} & 0.432 & 0.443 & 0.428 & 0.443\\\cline{2-15}

& \multirow{4}{*}{\rotatebox{90}{ETTm1}} & 96 & \textbf{0.312} & \textbf{0.348} & 0.319 & 0.353 & \underline{0.317} & \underline{0.352} & 0.321 & 0.355 & 0.318 & 0.354 & 0.319 & 0.354\\
& & 192 & 0.359 & \underline{0.375} & 0.361 & 0.378 & \textbf{0.357} & 0.375 & 0.372 & 0.385 & \underline{0.358} & \textbf{0.374} & 0.366 & 0.380\\
& & 336 & \underline{0.390} & \underline{0.397} & 0.393 & 0.401 & 0.391 & 0.400 & 0.407 & 0.409 & \textbf{0.390} & \textbf{0.396} & 0.396 & 0.402\\
& & 720 & \underline{0.450} & \underline{0.432} & 0.451 & 0.434 & \textbf{0.449} & 0.433 & 0.467 & 0.445 & 0.452 & \textbf{0.432} & 0.458 & 0.437\\\cline{2-15}

& \multirow{4}{*}{\rotatebox{90}{ETTm2}} & 96 & \textbf{0.167} & \textbf{0.250} & 0.168 & \underline{0.251} & 0.170 & 0.254 & 0.169 & 0.252 & 0.172 & 0.255 & \underline{0.168} & 0.252\\
& & 192 & \textbf{0.233} & \textbf{0.293} & 0.236 & 0.296 & 0.236 & 0.296 & 0.235 & \underline{0.295} & \underline{0.235} & 0.297 & 0.235 & 0.297\\
& & 336 & \textbf{0.293} & \textbf{0.333} & 0.296 & 0.334 & 0.295 & 0.334 & 0.297 & 0.335 & \underline{0.294} & \underline{0.334} & 0.299 & 0.338\\
& & 720 & \textbf{0.391} & \textbf{0.390} & 0.397 & 0.394 & 0.394 & 0.395 & 0.398 & 0.395 & \underline{0.394} & \underline{0.393} & 0.403 & 0.399\\\cline{2-15}

& \multirow{4}{*}{\rotatebox{90}{Exchange}} & 96 & \textbf{0.084} & \textbf{0.204} & 0.087 & 0.206 & \underline{0.086} & \underline{0.205} & 0.088 & 0.206 & 0.090 & 0.210 & 0.086 & 0.206\\
& & 192 & \textbf{0.182} & \textbf{0.303} & \underline{0.184} & \underline{0.304} & 0.187 & 0.306 & 0.187 & 0.306 & 0.191 & 0.310 & 0.188 & 0.308\\
& & 336 & \underline{0.345} & \textbf{0.423} & 0.353 & 0.429 & \textbf{0.344} & \underline{0.424} & 0.350 & 0.428 & 0.359 & 0.434 & 0.353 & 0.429\\
& & 720 & \textbf{0.842} & \textbf{0.688} & 0.886 & 0.706 & 0.893 & 0.711 & \underline{0.861} & \underline{0.698} & 0.883 & 0.707 & 0.900 & 0.716\\\cline{2-15}

& \multirow{4}{*}{\rotatebox{90}{Weather}} & 96 & \underline{0.151} & \underline{0.195} & 0.157 & 0.200 & 0.155 & 0.198 & 0.155 & 0.198 & 0.161 & 0.203 & \textbf{0.151} & \textbf{0.194}\\
& & 192 & \textbf{0.201} & \textbf{0.241} & 0.204 & 0.245 & 0.203 & 0.243 & 0.201 & 0.242 & 0.202 & 0.244 & \underline{0.201} & \underline{0.241}\\
& & 336 & \textbf{0.259} & \underline{0.285} & 0.263 & 0.287 & 0.262 & 0.286 & 0.263 & 0.288 & 0.262 & 0.286 & \underline{0.261} & \textbf{0.285}\\
& & 720 & \textbf{0.339} & \textbf{0.337} & \underline{0.340} & \underline{0.338} & 0.341 & 0.338 & 0.343 & 0.341 & 0.343 & 0.340 & 0.346 & 0.343\\\hline

\multicolumn{3}{c}{\# Top-2}
& \multicolumn{2}{c}{46}
& \multicolumn{2}{c}{9}
& \multicolumn{2}{c}{8}
& \multicolumn{2}{c}{6}
& \multicolumn{2}{c}{19}
& \multicolumn{2}{c}{8}
\\\hline

\end{tabular}}
\label{complete_results_timexer}
\end{table}

\begin{table}[h]
\caption{Complete results based on \textit{iTransformer} backbone.}
\centering
\resizebox{1\columnwidth}{!}{
\begin{tabular}{ccccccccccccccc}
\hline
\multirow{30}{*}{\rotatebox{90}{iTransformer}}&\multirow{2}{*}{D} & \multirow{2}{*}{PL} & \multicolumn{2}{c}{SGA} & \multicolumn{2}{c}{SA} & \multicolumn{2}{c}{Geometry} & \multicolumn{2}{c}{ProbSparse} & \multicolumn{2}{c}{AutoCorrelation} & \multicolumn{2}{c}{TSSA}\\
&&& MSE & MAE & MSE & MAE & MSE & MAE & MSE & MAE & MSE & MAE & MSE & MAE\\\cline{2-15}

& \multirow{4}{*}{\rotatebox{90}{ETTh1}}& 96 & 0.389 & 0.404 & \textbf{0.385} & \underline{0.401} & 0.393 & 0.406 & 0.391 & 0.404 & \underline{0.388} & 0.402 & 0.389 & \textbf{0.400}\\
& & 192 & 0.443 & 0.434 & 0.440 & 0.432 & 0.447 & 0.437 & 0.445 & 0.435 & \textbf{0.436} & \textbf{0.427} & \underline{0.437} & \underline{0.428}\\
& & 336 & \textbf{0.480} & \underline{0.451} & 0.484 & 0.455 & 0.494 & 0.461 & 0.495 & 0.459 & 0.487 & 0.454 & \underline{0.482} & \textbf{0.450}\\
& & 720 & \textbf{0.490} & \textbf{0.474} & 0.498 & 0.485 & 0.534 & 0.502 & 0.573 & 0.528 & 0.512 & 0.491 & \underline{0.492} & \underline{0.476}\\
\cline{2-15}

& \multirow{4}{*}{\rotatebox{90}{ETTh2}}& 96 & 0.298 & 0.348 & 0.295 & 0.345 & 0.295 & 0.345 & \textbf{0.290} & \textbf{0.342} & \underline{0.292} & \underline{0.342} & 0.292 & 0.342\\
& & 192 & 0.379 & 0.396 & 0.376 & 0.394 & 0.378 & 0.396 & 0.379 & 0.397 & \textbf{0.371} & \textbf{0.392} & \underline{0.374} & \underline{0.394}\\
& & 336 & 0.419 & 0.429 & 0.421 & 0.428 & 0.425 & 0.430 & \underline{0.418} & 0.430 & 0.419 & \underline{0.428} & \textbf{0.418} & \textbf{0.428}\\
& & 720 & \underline{0.423} & 0.442 & \textbf{0.423} & \textbf{0.441} & 0.433 & 0.446 & 0.428 & 0.444 & 0.423 & \underline{0.442} & 0.426 & 0.442\\
\cline{2-15}

& \multirow{4}{*}{\rotatebox{90}{ETTm1}}& 96 & \textbf{0.322} & \textbf{0.359} & 0.331 & 0.364 & 0.331 & 0.366 & 0.324 & 0.361 & 0.349 & 0.374 & \underline{0.324} & \underline{0.361}\\
& & 192 & \underline{0.369} & \textbf{0.381} & 0.377 & 0.386 & 0.376 & 0.386 & 0.370 & 0.383 & 0.382 & 0.388 & \textbf{0.369} & \underline{0.381}\\
& & 336 & \textbf{0.404} & \underline{0.405} & 0.413 & 0.410 & 0.418 & 0.412 & 0.409 & 0.408 & 0.422 & 0.415 & \underline{0.405} & \textbf{0.404}\\
& & 720 & \textbf{0.469} & \underline{0.441} & 0.484 & 0.449 & 0.486 & 0.450 & 0.477 & 0.445 & 0.474 & 0.441 & \underline{0.471} & \textbf{0.440}\\
\cline{2-15}

& \multirow{4}{*}{\rotatebox{90}{ETTm2}}& 96 & \underline{0.176} & \underline{0.259} & 0.180 & 0.262 & 0.180 & 0.263 & 0.176 & 0.259 & 0.178 & 0.261 & \textbf{0.174} & \textbf{0.257}\\
& & 192 & \underline{0.243} & \underline{0.302} & 0.246 & 0.304 & 0.247 & 0.306 & \textbf{0.239} & \textbf{0.300} & 0.243 & 0.303 & 0.243 & 0.303\\
& & 336 & 0.305 & 0.342 & 0.311 & 0.346 & 0.310 & 0.345 & \textbf{0.303} & \textbf{0.340} & \underline{0.304} & \underline{0.341} & 0.307 & 0.342\\
& & 720 & 0.408 & 0.401 & 0.410 & 0.402 & 0.411 & 0.402 & \textbf{0.405} & \textbf{0.398} & \underline{0.407} & \underline{0.400} & 0.408 & 0.400\\
\cline{2-15}

& \multirow{4}{*}{\rotatebox{90}{Exchange}}& 96 & \underline{0.087} & 0.208 & 0.087 & 0.207 & 0.087 & 0.209 & 0.088 & 0.209 & 0.087 & \underline{0.206} & \textbf{0.086} & \textbf{0.205}\\
& & 192 & \textbf{0.181} & \textbf{0.304} & 0.182 & 0.304 & 0.182 & 0.306 & 0.183 & 0.305 & 0.184 & 0.304 & \underline{0.182} & \underline{0.304}\\
& & 336 & \textbf{0.331} & \textbf{0.416} & 0.336 & 0.421 & \underline{0.333} & \underline{0.418} & 0.341 & 0.423 & 0.338 & 0.422 & 0.336 & 0.420\\
& & 720 & \textbf{0.844} & \textbf{0.694} & 0.861 & \underline{0.699} & \underline{0.856} & 0.700 & 0.884 & 0.716 & 0.878 & 0.707 & 0.860 & 0.700\\
\cline{2-15}

& \multirow{4}{*}{\rotatebox{90}{Weather}}& 96 & \underline{0.157} & \underline{0.200} & 0.169 & 0.207 & 0.172 & 0.209 & \textbf{0.154} & \textbf{0.197} & 0.176 & 0.215 & 0.157 & 0.202\\
& & 192 & \underline{0.207} & \underline{0.246} & 0.221 & 0.253 & 0.222 & 0.254 & \textbf{0.206} & \textbf{0.245} & 0.223 & 0.256 & 0.207 & 0.247\\
& & 336 & \textbf{0.265} & \underline{0.290} & 0.280 & 0.296 & 0.280 & 0.296 & \underline{0.266} & 0.290 & 0.282 & 0.299 & 0.267 & \textbf{0.290}\\
& & 720 & \underline{0.348} & 0.346 & 0.361 & 0.351 & 0.359 & 0.348 & \textbf{0.348} & \textbf{0.343} & 0.358 & 0.347 & 0.351 & \underline{0.345}\\\hline

\multicolumn{3}{c}{\# Top-2}
& \multicolumn{2}{c}{31}
& \multicolumn{2}{c}{5}
& \multicolumn{2}{c}{3}
& \multicolumn{2}{c}{16}
& \multicolumn{2}{c}{14}
& \multicolumn{2}{c}{27}
\\\hline

\end{tabular}}
\label{complete_results_itransformer}
\end{table}

\begin{table}[h!]
\caption{Complete results based on \textit{SimpleTM} backbone.}
\centering
\resizebox{1\columnwidth}{!}{
\begin{tabular}{ccccccccccccccc}
\hline
\multirow{30}{*}{\rotatebox{90}{SimpleTM}}&\multirow{2}{*}{D} & \multirow{2}{*}{PL} & \multicolumn{2}{c}{SGA} & \multicolumn{2}{c}{SA} & \multicolumn{2}{c}{Geometry} & \multicolumn{2}{c}{ProbSparse} & \multicolumn{2}{c}{AutoCorrelation} & \multicolumn{2}{c}{TSSA}\\
&&& MSE& MAE & MSE & MAE & MSE & MAE & MSE & MAE & MSE & MAE & MSE & MAE\\\cline{2-15}
& \multirow{4}{*}{\rotatebox{90}{ETTh1}}& 96 & \textbf{0.378} & \underline{0.396} & 0.382 & 0.399 & \underline{0.379} & \textbf{0.395} & 0.379 & 0.396 & 0.381 & 0.397 & 0.383 & 0.399\\
& & 192 & \textbf{0.424} & \textbf{0.422} & 0.429 & 0.425 & \underline{0.425} & \underline{0.422} & 0.426 & 0.423 & 0.427 & 0.423 & 0.427 & 0.423\\
& & 336 & \textbf{0.469} & \textbf{0.447} & 0.477 & 0.454 & \underline{0.471} & \underline{0.449} & 0.473 & 0.452 & 0.489 & 0.461 & 0.480 & 0.455\\
& & 720 & \textbf{0.466} & \textbf{0.461} & 0.480 & 0.474 & 0.475 & 0.468 & \underline{0.471} & 0.470 & 0.491 & 0.479 & 0.475 & \underline{0.465}\\\cline{2-15}

& \multirow{4}{*}{\rotatebox{90}{ETTh2}}& 96 & \textbf{0.288} & \textbf{0.338} & 0.292 & 0.342 & \underline{0.289} & \underline{0.339} & 0.295 & 0.344 & 0.297 & 0.346 & 0.292 & 0.342\\
& & 192 & \textbf{0.364} & \underline{0.388} & 0.373 & 0.393 & \underline{0.365} & \textbf{0.387} & 0.374 & 0.392 & 0.371 & 0.391 & 0.374 & 0.392\\
& & 336 & \textbf{0.415} & \textbf{0.426} & 0.419 & 0.428 & 0.417 & \underline{0.426} & \underline{0.416} & 0.427 & 0.425 & 0.430 & 0.417 & 0.427\\
& & 720 & \textbf{0.416} & \textbf{0.438} & \underline{0.419} & 0.439 & 0.424 & 0.441 & 0.421 & \underline{0.439} & 0.435 & 0.447 & 0.426 & 0.442\\\cline{2-15}

& \multirow{4}{*}{\rotatebox{90}{ETTm1}}& 96 & \textbf{0.315} & \textbf{0.352} & 0.320 & 0.355 & 0.320 & 0.354 & \underline{0.317} & \underline{0.353} & 0.322 & 0.356 & 0.335 & 0.368\\
& & 192 & \textbf{0.359} & \textbf{0.374} & 0.361 & 0.376 & 0.359 & 0.375 & \underline{0.359} & \underline{0.374} & 0.377 & 0.388 & 0.361 & 0.376\\
& & 336 & \textbf{0.390} & \textbf{0.396} & 0.392 & 0.397 & \underline{0.390} & \underline{0.396} & 0.391 & 0.397 & 0.409 & 0.405 & 0.399 & 0.401\\
& & 720 & \textbf{0.454} & \textbf{0.432} & 0.457 & 0.434 & \underline{0.455} & \underline{0.432} & 0.456 & 0.433 & 0.458 & 0.435 & 0.460 & 0.435\\\cline{2-15}

& \multirow{4}{*}{\rotatebox{90}{ETTm2}}& 96 & \textbf{0.172} & \textbf{0.253} & 0.174 & 0.256 & 0.173 & 0.255 & \underline{0.173} & \underline{0.255} & 0.176 & 0.258 & 0.175 & 0.256\\
& & 192 & \textbf{0.235} & \textbf{0.295} & 0.238 & 0.298 & \underline{0.235} & \underline{0.296} & 0.238 & 0.297 & 0.247 & 0.306 & 0.238 & 0.297\\
& & 336 & \textbf{0.293} & \textbf{0.333} & 0.295 & 0.334 & \underline{0.293} & \underline{0.333} & 0.295 & 0.334 & 0.303 & 0.341 & 0.295 & 0.334\\
& & 720 & \textbf{0.391} & \textbf{0.391} & 0.395 & 0.394 & 0.394 & \underline{0.392} & 0.395 & 0.393 & \underline{0.392} & 0.392 & 0.397 & 0.394\\\cline{2-15}

& \multirow{4}{*}{\rotatebox{90}{Exchange}}& 96 & \underline{0.090} & \underline{0.209} & 0.097 & 0.221 & 0.092 & 0.210 & 0.096 & 0.216 & \textbf{0.089} & \textbf{0.209} & 0.093 & 0.213\\
& & 192 & \underline{0.185} & \textbf{0.306} & 0.191 & 0.313 & 0.187 & 0.309 & 0.192 & 0.312 & 0.192 & 0.314 & \textbf{0.184} & \underline{0.307}\\
& & 336 & \textbf{0.343} & \textbf{0.426} & \underline{0.348} & \underline{0.429} & 0.358 & 0.435 & 0.384 & 0.453 & 0.366 & 0.441 & 0.369 & 0.440\\
& & 720 & \textbf{0.882} & \textbf{0.710} & 0.917 & 0.725 & \underline{0.894} & \underline{0.714} & 0.931 & 0.730 & 0.936 & 0.732 & 0.922 & 0.729\\\cline{2-15}

& \multirow{4}{*}{\rotatebox{90}{Weather}}& 96 & \underline{0.162} & \underline{0.204} & 0.164 & 0.206 & \textbf{0.152} & \textbf{0.195} & 0.163 & 0.205 & 0.195 & 0.235 & 0.174 & 0.220\\
& & 192 & \underline{0.208} & \underline{0.247} & 0.209 & 0.247 & \textbf{0.202} & \textbf{0.242} & 0.210 & 0.248 & 0.244 & 0.272 & 0.217 & 0.256\\
& & 336 & \textbf{0.264} & \textbf{0.287} & \underline{0.264} & \underline{0.288} & 0.265 & 0.288 & 0.265 & 0.288 & 0.272 & 0.294 & 0.267 & 0.290\\
& & 720 & \textbf{0.344} & \textbf{0.340} & \underline{0.345} & \underline{0.340} & 0.347 & 0.342 & 0.345 & 0.340 & 0.367 & 0.355 & 0.345 & 0.341\\\hline

\multicolumn{3}{c}{\# Top-2}
& \multicolumn{2}{c}{48}
& \multicolumn{2}{c}{7}
& \multicolumn{2}{c}{26}
& \multicolumn{2}{c}{9}
& \multicolumn{2}{c}{3}
& \multicolumn{2}{c}{3}
\\\hline

\end{tabular}}
\label{complete_results_simpleTM}
\end{table}

\begin{table}[t]
\caption{Complete results based on \textit{CARD} backbone.}
\centering
\resizebox{1\columnwidth}{!}{
\begin{tabular}{ccccccccccccccc}
\hline
\multirow{30}{*}{\rotatebox{90}{CARD}}&\multirow{2}{*}{D} & \multirow{2}{*}{PL} & \multicolumn{2}{c}{SGA} & \multicolumn{2}{c}{SA} & \multicolumn{2}{c}{Geometry} & \multicolumn{2}{c}{ProbSparse} & \multicolumn{2}{c}{AutoCorrelation} & \multicolumn{2}{c}{TSSA}\\
&&& MSE & MAE & MSE & MAE & MSE & MAE & MSE & MAE & MSE & MAE & MSE & MAE\\\cline{2-15}

& \multirow{4}{*}{\rotatebox{90}{ETTh1}}& 96 & \textbf{0.386} & \underline{0.395} & 0.387 & 0.397 & 0.387 & 0.395 & \underline{0.387} & \textbf{0.394} & 0.387 & 0.395 & 0.387 & 0.395\\
& & 192 & \textbf{0.436} & \textbf{0.423} & 0.445 & 0.432 & 0.440 & 0.427 & 0.441 & 0.427 & 0.440 & 0.426 & \underline{0.437} & \underline{0.424}\\
& & 336 & \underline{0.480} & \underline{0.445} & 0.483 & 0.450 & 0.484 & 0.449 & 0.481 & 0.446 & 0.481 & 0.447 & \textbf{0.476} & \textbf{0.442}\\
& & 720 & \underline{0.477} & 0.466 & 0.484 & 0.472 & 0.482 & 0.470 & 0.477 & \underline{0.465} & 0.481 & 0.470 & \textbf{0.476} & \textbf{0.461}\\\cline{2-15}

& \multirow{4}{*}{\rotatebox{90}{ETTh2}}& 96 & \underline{0.284} & \underline{0.336} & \textbf{0.282} & \textbf{0.335} & 0.289 & 0.337 & 0.287 & 0.336 & 0.290 & 0.337 & 0.290 & 0.338\\
& & 192 & \textbf{0.361} & \underline{0.386} & 0.366 & 0.386 & \underline{0.365} & \textbf{0.386} & 0.370 & 0.388 & 0.377 & 0.390 & 0.377 & 0.390\\
& & 336 & \textbf{0.405} & \textbf{0.419} & 0.411 & 0.421 & \underline{0.410} & \underline{0.421} & 0.417 & 0.426 & 0.418 & 0.424 & 0.420 & 0.427\\
& & 720 & \textbf{0.406} & \textbf{0.431} & 0.419 & 0.437 & \underline{0.417} & \underline{0.436} & 0.420 & 0.439 & 0.418 & 0.437 & 0.436 & 0.447\\\cline{2-15}

& \multirow{4}{*}{\rotatebox{90}{ETTm1}}& 96 & \underline{0.331} & 0.364 & \textbf{0.331} & 0.364 & 0.332 & \textbf{0.359} & 0.332 & \underline{0.363} & 0.351 & 0.371 & 0.337 & 0.365\\
& & 192 & 0.372 & 0.384 & \textbf{0.363} & \textbf{0.378} & 0.377 & 0.381 & \underline{0.371} & 0.381 & 0.380 & 0.383 & 0.372 & \underline{0.380}\\
& & 336 & \underline{0.401} & 0.401 & \textbf{0.394} & \underline{0.399} & 0.409 & 0.402 & 0.406 & 0.403 & 0.410 & 0.401 & 0.402 & \textbf{0.399}\\
& & 720 & \underline{0.464} & \underline{0.435} & \textbf{0.461} & 0.436 & 0.471 & 0.437 & 0.469 & 0.438 & 0.475 & 0.439 & 0.466 & \textbf{0.434}\\\cline{2-15}

& \multirow{4}{*}{\rotatebox{90}{ETTm2}}& 96 & \underline{0.171} & \underline{0.255} & \textbf{0.169} & \textbf{0.253} & 0.178 & 0.261 & 0.176 & 0.259 & 0.180 & 0.261 & 0.174 & 0.256\\
& & 192 & \underline{0.235} & \underline{0.297} & \textbf{0.231} & \textbf{0.293} & 0.241 & 0.300 & 0.239 & 0.300 & 0.245 & 0.303 & 0.240 & 0.298\\
& & 336 & \underline{0.295} & \underline{0.335} & \textbf{0.292} & \textbf{0.333} & 0.301 & 0.337 & 0.301 & 0.338 & 0.304 & 0.339 & 0.300 & 0.336\\
& & 720 & \textbf{0.391} & \textbf{0.391} & \underline{0.394} & \underline{0.393} & 0.401 & 0.394 & 0.403 & 0.396 & 0.406 & 0.396 & 0.402 & 0.394\\\cline{2-15}

& \multirow{4}{*}{\rotatebox{90}{Exchange}}& 96 & \textbf{0.084} & 0.202 & 0.086 & 0.206 & 0.086 & 0.205 & 0.088 & 0.208 & 0.086 & \underline{0.202} & \underline{0.085} & \textbf{0.202}\\
& & 192 & 0.174 & 0.296 & 0.180 & 0.303 & \textbf{0.173} & \underline{0.296} & 0.177 & 0.298 & \underline{0.174} & \textbf{0.295} & 0.185 & 0.304\\
& & 336 & \underline{0.332} & \underline{0.414} & \textbf{0.325} & \textbf{0.412} & 0.361 & 0.432 & 0.393 & 0.446 & 0.333 & 0.416 & 0.366 & 0.435\\
& & 720 & \textbf{0.819} & \textbf{0.681} & 1.319 & 0.847 & \underline{0.848} & \underline{0.694} & 1.110 & 0.783 & 1.093 & 0.790 & 0.971 & 0.737\\\cline{2-15}

& \multirow{4}{*}{\rotatebox{90}{Weather}}& 96 & \underline{0.158} & \underline{0.204} & \textbf{0.158} & \textbf{0.202} & 0.173 & 0.212 & 0.159 & 0.205 & 0.185 & 0.222 & 0.160 & 0.205\\
& & 192 & 0.209 & 0.250 & \textbf{0.208} & \underline{0.249} & 0.223 & 0.258 & 0.212 & 0.253 & 0.228 & 0.258 & \underline{0.208} & \textbf{0.248}\\
& & 336 & \underline{0.267} & \underline{0.292} & \textbf{0.266} & \textbf{0.291} & 0.279 & 0.299 & 0.277 & 0.300 & 0.285 & 0.299 & 0.270 & 0.293\\
& & 720 & \textbf{0.346} & \underline{0.344} & 0.350 & 0.346 & 0.355 & 0.349 & 0.364 & 0.352 & 0.360 & 0.347 & \underline{0.348} & \textbf{0.344}\\\hline

\multicolumn{3}{c}{\# Top-2}
& \multicolumn{2}{c}{38}
& \multicolumn{2}{c}{24}
& \multicolumn{2}{c}{11}
& \multicolumn{2}{c}{5}
& \multicolumn{2}{c}{3}
& \multicolumn{2}{c}{15}
\\\hline

\end{tabular}}
\label{complete_results_card}
\end{table}

\begin{table}[h!]
\caption{Complete results based on \textit{FEDformer} backbone.}
\centering
\resizebox{1\columnwidth}{!}{
\begin{tabular}{ccccccccccccccc}
\hline
\multirow{30}{*}{\rotatebox{90}{FEDformer}}&\multirow{2}{*}{D} & \multirow{2}{*}{PL} & \multicolumn{2}{c}{SGA} & \multicolumn{2}{c}{SA} & \multicolumn{2}{c}{Geometry} & \multicolumn{2}{c}{ProbSparse} & \multicolumn{2}{c}{AutoCorrelation} & \multicolumn{2}{c}{TSSA}\\
&&& MSE & MAE & MSE & MAE & MSE & MAE & MSE & MAE & MSE & MAE & MSE & MAE\\\cline{2-15}

& \multirow{4}{*}{\rotatebox{90}{ETTh1}}& 96 & 0.405 & 0.441 & 0.486 & 0.487 & 0.508 & 0.508 & 0.549 & 0.511 & \textbf{0.375} & \textbf{0.411} & \underline{0.390} & \underline{0.416}\\
& & 192 & \underline{0.444} & 0.461 & 0.548 & 0.524 & 0.561 & 0.537 & 0.617 & 0.534 & \textbf{0.424} & \textbf{0.442} & 0.451 & \underline{0.457}\\
& & 336 & \underline{0.481} & \underline{0.480} & 0.690 & 0.624 & 0.570 & 0.546 & 0.669 & 0.582 & \textbf{0.471} & \textbf{0.464} & 0.606 & 0.556\\
& & 720 & \textbf{0.494} & \underline{0.508} & 0.652 & 0.605 & 0.596 & 0.567 & 0.734 & 0.633 & \underline{0.500} & \textbf{0.499} & 0.716 & 0.626\\\cline{2-15}

& \multirow{4}{*}{\rotatebox{90}{ETTh2}}& 96 & 0.348 & 0.399 & 0.400 & 0.426 & 0.404 & 0.443 & 0.411 & 0.430 & \underline{0.338} & \underline{0.382} & \textbf{0.334} & \textbf{0.381}\\
& & 192 & \underline{0.425} & \underline{0.437} & 0.459 & 0.455 & 0.493 & 0.485 & 1.618 & 0.922 & \textbf{0.418} & \textbf{0.428} & 0.679 & 0.559\\
& & 336 & \textbf{0.448} & \underline{0.461} & 0.524 & 0.503 & 0.513 & 0.502 & 1.217 & 0.811 & \underline{0.453} & \textbf{0.460} & 0.629 & 0.542\\
& & 720 & \textbf{0.433} & \textbf{0.460} & 0.542 & 0.529 & 0.491 & 0.519 & 0.922 & 0.702 & \underline{0.452} & \underline{0.464} & 0.888 & 0.667\\\cline{2-15}

& \multirow{4}{*}{\rotatebox{90}{ETTm1}}& 96 & \textbf{0.329} & \textbf{0.386} & 0.421 & 0.441 & 0.443 & 0.453 & 0.499 & 0.489 & \underline{0.363} & \underline{0.406} & 0.465 & 0.470\\
& & 192 & \textbf{0.376} & \textbf{0.413} & 0.479 & 0.473 & 0.566 & 0.520 & 0.534 & 0.504 & \underline{0.430} & \underline{0.437} & 0.557 & 0.531\\
& & 336 & \textbf{0.407} & \textbf{0.436} & 0.543 & 0.517 & 0.625 & 0.546 & 0.665 & 0.574 & \underline{0.447} & \underline{0.449} & 0.607 & 0.544\\
& & 720 & \textbf{0.476} & \underline{0.485} & 0.654 & 0.588 & 0.783 & 0.635 & 0.787 & 0.625 & \underline{0.518} & \textbf{0.481} & 0.793 & 0.661\\\cline{2-15}

& \multirow{4}{*}{\rotatebox{90}{ETTm2}}& 96 & \textbf{0.182} & \textbf{0.275} & 0.201 & 0.290 & 0.202 & 0.291 & 0.254 & 0.334 & \underline{0.192} & \underline{0.279} & 0.315 & 0.398\\
& & 192 & \textbf{0.249} & \underline{0.319} & 0.282 & 0.341 & 0.274 & 0.337 & 0.268 & 0.336 & \underline{0.256} & \textbf{0.317} & 0.460 & 0.477\\
& & 336 & \underline{0.319} & \underline{0.367} & 0.344 & 0.380 & 0.387 & 0.416 & 0.331 & 0.372 & \textbf{0.315} & \textbf{0.355} & 0.689 & 0.621\\
& & 720 & \underline{0.438} & \underline{0.439} & 0.469 & 0.448 & 0.555 & 0.506 & 2.175 & 0.996 & \textbf{0.417} & \textbf{0.413} & 2.565 & 1.190\\\cline{2-15}

& \multirow{4}{*}{\rotatebox{90}{Exchange}}& 96 & \textbf{0.134} & \textbf{0.265} & 0.134 & \underline{0.266} & \underline{0.134} & 0.268 & 0.174 & 0.314 & 0.162 & 0.291 & 0.151 & 0.284\\
& & 192 & \underline{0.239} & \underline{0.357} & \textbf{0.223} & \textbf{0.349} & 0.299 & 0.402 & 0.339 & 0.430 & 0.287 & 0.387 & 0.379 & 0.451\\
& & 336 & \textbf{0.427} & \textbf{0.481} & 0.510 & 0.533 & 0.486 & 0.522 & \underline{0.441} & \underline{0.484} & 0.450 & 0.492 & 0.554 & 0.549\\
& & 720 & \underline{1.064} & \underline{0.789} & 1.318 & 0.890 & \textbf{0.986} & \textbf{0.769} & 1.483 & 0.931 & 1.198 & 0.841 & 1.269 & 0.864\\\cline{2-15}

& \multirow{4}{*}{\rotatebox{90}{Weather}}& 96 & 0.228 & 0.299 & \textbf{0.216} & \textbf{0.285} & 0.240 & 0.305 & 0.429 & 0.453 & \underline{0.225} & \underline{0.298} & 0.298 & 0.369\\
& & 192 & \underline{0.303} & \underline{0.368} & 0.321 & 0.368 & 0.338 & 0.388 & 0.491 & 0.495 & \textbf{0.263} & \textbf{0.324} & 0.387 & 0.429\\
& & 336 & 0.378 & 0.417 & 0.433 & 0.452 & \underline{0.377} & \underline{0.414} & 0.798 & 0.638 & \textbf{0.331} & \textbf{0.371} & 0.520 & 0.499\\
& & 720 & 0.476 & 0.456 & 0.525 & 0.490 & \underline{0.456} & \underline{0.454} & 1.061 & 0.749 & \textbf{0.404} & \textbf{0.412} & 0.879 & 0.686\\\hline

\multicolumn{3}{c}{\# Top-2}
& \multicolumn{2}{c}{37}
& \multicolumn{2}{c}{5}
& \multicolumn{2}{c}{7}
& \multicolumn{2}{c}{2}
& \multicolumn{2}{c}{40}
& \multicolumn{2}{c}{5}
\\\hline

\end{tabular}}
\label{complete_results_fedformer}
\end{table}

\begin{table}[t]
\caption{Complete results based on \textit{PAttn} backbone.}
\centering
\resizebox{1\columnwidth}{!}{
\begin{tabular}{ccccccccccccccc}
\hline
\multirow{30}{*}{\rotatebox{90}{PAttn}}&\multirow{2}{*}{D} & \multirow{2}{*}{PL} & \multicolumn{2}{c}{SGA} & \multicolumn{2}{c}{SA} & \multicolumn{2}{c}{Geometry} & \multicolumn{2}{c}{ProbSparse} & \multicolumn{2}{c}{AutoCorrelation} & \multicolumn{2}{c}{TSSA}\\
&&& MSE & MAE & MSE & MAE & MSE & MAE & MSE & MAE & MSE & MAE & MSE & MAE\\\cline{2-15}

& \multirow{4}{*}{\rotatebox{90}{ETTh1}}& 96 & \textbf{0.385} & \underline{0.393} & 0.385 & \textbf{0.392} & 0.386 & 0.396 & 0.390 & 0.395 & 0.389 & 0.397 & \underline{0.385} & 0.394\\
& & 192 & \underline{0.436} & \underline{0.425} & 0.437 & \textbf{0.424} & \textbf{0.436} & 0.427 & 0.453 & 0.432 & 0.441 & 0.425 & 0.441 & 0.427\\
& & 336 & \underline{0.487} & 0.454 & 0.488 & 0.453 & 0.488 & 0.456 & 0.489 & \underline{0.452} & \textbf{0.484} & \textbf{0.449} & 0.490 & 0.456\\
& & 720 & \underline{0.504} & \underline{0.480} & 0.540 & 0.501 & 0.536 & 0.497 & 0.515 & 0.487 & \textbf{0.494} & \textbf{0.478} & 0.506 & 0.483\\
\cline{2-15}

& \multirow{4}{*}{\rotatebox{90}{ETTh2}}& 96 & \textbf{0.282} & \textbf{0.333} & \underline{0.284} & \underline{0.335} & 0.287 & 0.339 & 0.285 & 0.336 & 0.288 & 0.337 & 0.292 & 0.339\\
& & 192 & \textbf{0.360} & \textbf{0.382} & \underline{0.363} & 0.387 & 0.364 & 0.390 & 0.364 & \underline{0.387} & 0.370 & 0.389 & 0.366 & 0.388\\
& & 336 & \textbf{0.405} & \textbf{0.418} & 0.413 & 0.426 & \underline{0.408} & 0.426 & 0.411 & 0.423 & 0.412 & 0.424 & 0.413 & \underline{0.423}\\
& & 720 & \textbf{0.415} & \underline{0.437} & 0.428 & 0.443 & 0.423 & 0.444 & 0.421 & 0.442 & \underline{0.417} & \textbf{0.437} & 0.424 & 0.442\\
\cline{2-15}

& \multirow{4}{*}{\rotatebox{90}{ETTm1}}& 96 & \textbf{0.315} & \underline{0.354} & 0.326 & 0.360 & \underline{0.316} & 0.354 & 0.320 & 0.356 & 0.332 & 0.365 & 0.318 & \textbf{0.354}\\
& & 192 & \textbf{0.359} & \underline{0.378} & 0.364 & 0.378 & \underline{0.359} & \textbf{0.377} & 0.366 & 0.378 & 0.373 & 0.381 & 0.366 & 0.379\\
& & 336 & \textbf{0.388} & 0.399 & 0.391 & \textbf{0.398} & \underline{0.390} & \underline{0.399} & 0.394 & 0.399 & 0.400 & 0.401 & 0.396 & 0.399\\
& & 720 & \textbf{0.451} & 0.434 & 0.455 & \textbf{0.433} & \underline{0.454} & 0.435 & 0.456 & \underline{0.434} & 0.467 & 0.437 & 0.460 & 0.435\\
\cline{2-15}

& \multirow{4}{*}{\rotatebox{90}{ETTm2}}& 96 & \textbf{0.175} & \textbf{0.259} & 0.176 & 0.261 & 0.176 & 0.262 & \underline{0.175} & \underline{0.259} & 0.177 & 0.261 & 0.176 & 0.259\\
& & 192 & \textbf{0.241} & \textbf{0.302} & 0.246 & 0.307 & 0.245 & 0.308 & \underline{0.242} & \underline{0.303} & 0.243 & 0.304 & 0.244 & 0.306\\
& & 336 & \textbf{0.301} & \textbf{0.341} & 0.307 & 0.347 & 0.306 & 0.347 & \underline{0.305} & \underline{0.345} & 0.307 & 0.346 & 0.308 & 0.347\\
& & 720 & \textbf{0.401} & \textbf{0.398} & 0.410 & 0.405 & 0.411 & 0.407 & \underline{0.407} & \underline{0.405} & 0.410 & 0.406 & 0.409 & 0.405\\
\cline{2-15}

& \multirow{4}{*}{\rotatebox{90}{Exchange}}& 96 & \textbf{0.083} & \textbf{0.202} & 0.085 & 0.204 & \underline{0.085} & \underline{0.203} & 0.089 & 0.208 & 0.088 & 0.207 & 0.090 & 0.209\\
& & 192 & \textbf{0.176} & \textbf{0.299} & 0.190 & 0.309 & 0.187 & 0.307 & \underline{0.185} & \underline{0.305} & 0.186 & 0.306 & 0.185 & 0.306\\
& & 336 & \textbf{0.340} & \textbf{0.419} & 0.359 & 0.434 & 0.351 & 0.428 & 0.349 & 0.427 & \underline{0.346} & \underline{0.426} & 0.346 & 0.427\\
& & 720 & \textbf{0.851} & \textbf{0.690} & \underline{0.855} & \underline{0.696} & 0.883 & 0.705 & 0.869 & 0.703 & 0.870 & 0.703 & 0.858 & 0.697\\
\cline{2-15}

& \multirow{4}{*}{\rotatebox{90}{Weather}}& 96 & 0.179 & 0.217 & 0.179 & 0.217 & \underline{0.179} & 0.218 & 0.180 & \underline{0.217} & 0.181 & 0.218 & \textbf{0.178} & \textbf{0.215}\\
& & 192 & 0.225 & 0.257 & \underline{0.225} & 0.256 & 0.225 & 0.257 & 0.226 & \underline{0.256} & 0.227 & 0.258 & \textbf{0.224} & \textbf{0.255}\\
& & 336 & 0.279 & 0.295 & \underline{0.279} & 0.296 & 0.280 & 0.297 & 0.280 & \underline{0.295} & 0.279 & 0.296 & \textbf{0.278} & \textbf{0.294}\\
& & 720 & \textbf{0.355} & \underline{0.345} & 0.355 & 0.345 & \underline{0.355} & 0.346 & 0.355 & \textbf{0.345} & 0.356 & 0.345 & 0.355 & 0.345\\
\hline

\multicolumn{3}{c}{\# Top-2}
& \multicolumn{2}{c}{39}
& \multicolumn{2}{c}{11}
& \multicolumn{2}{c}{12}
& \multicolumn{2}{c}{17}
& \multicolumn{2}{c}{8}
& \multicolumn{2}{c}{9}
\\\hline

\end{tabular}}
\label{complete_results_pattn}
\end{table}

\begin{table}[h!]
\caption{Complete results based on \textit{MultiPatchFormer} backbone.}
\centering
\resizebox{1\columnwidth}{!}{
\begin{tabular}{ccccccccccccccc}
\hline
\multirow{30}{*}{\rotatebox{90}{MultiPatchFormer}}&\multirow{2}{*}{D} & \multirow{2}{*}{PL} & \multicolumn{2}{c}{SGA} & \multicolumn{2}{c}{SA} & \multicolumn{2}{c}{Geometry} & \multicolumn{2}{c}{ProbSparse} & \multicolumn{2}{c}{AutoCorrelation} & \multicolumn{2}{c}{TSSA}\\
&&& MSE & MAE & MSE & MAE & MSE & MAE & MSE & MAE & MSE & MAE & MSE & MAE\\\cline{2-15}

& \multirow{4}{*}{\rotatebox{90}{ETTh1}}& 96 & \underline{0.377} & \underline{0.391} & \textbf{0.376} & 0.392 & 0.380 & 0.396 & 0.383 & 0.397 & 0.377 & \textbf{0.391} & 0.381 & 0.393\\
& & 192 & \underline{0.428} & \textbf{0.421} & \textbf{0.427} & \underline{0.421} & 0.433 & 0.426 & 0.435 & 0.428 & 0.430 & 0.422 & 0.434 & 0.423\\
& & 336 & \underline{0.468} & \textbf{0.440} & 0.469 & \underline{0.442} & \textbf{0.468} & 0.446 & 0.481 & 0.452 & 0.471 & 0.443 & 0.475 & 0.443\\
& & 720 & \textbf{0.476} & \textbf{0.463} & 0.492 & 0.476 & 0.495 & 0.479 & 0.489 & 0.473 & \underline{0.485} & \underline{0.471} & 0.491 & 0.471\\\cline{2-15}

& \multirow{4}{*}{\rotatebox{90}{ETTh2}}& 96 & \textbf{0.284} & \textbf{0.334} & 0.286 & 0.336 & \underline{0.285} & 0.336 & 0.290 & 0.340 & 0.288 & 0.336 & 0.287 & \underline{0.335}\\
& & 192 & \textbf{0.362} & \textbf{0.383} & 0.366 & 0.387 & \underline{0.363} & \underline{0.386} & 0.366 & 0.389 & 0.368 & 0.389 & 0.370 & 0.388\\
& & 336 & \underline{0.408} & \textbf{0.420} & 0.408 & \underline{0.420} & \textbf{0.408} & 0.421 & 0.412 & 0.426 & 0.412 & 0.421 & 0.417 & 0.426\\
& & 720 & \textbf{0.415} & \textbf{0.435} & \underline{0.417} & 0.439 & 0.423 & 0.441 & 0.443 & 0.451 & 0.418 & \underline{0.436} & 0.428 & 0.444\\\cline{2-15}

& \multirow{4}{*}{\rotatebox{90}{ETTm1}}& 96 & \underline{0.312} & \underline{0.350} & 0.315 & 0.351 & \textbf{0.312} & \textbf{0.350} & 0.323 & 0.359 & 0.325 & 0.358 & 0.318 & 0.355\\
& & 192 & \textbf{0.358} & \textbf{0.375} & 0.363 & 0.378 & \underline{0.359} & \underline{0.377} & 0.370 & 0.384 & 0.371 & 0.382 & 0.363 & 0.377\\
& & 336 & \textbf{0.390} & \underline{0.399} & \underline{0.394} & 0.400 & 0.395 & 0.402 & 0.408 & 0.409 & 0.401 & 0.402 & 0.394 & \textbf{0.399}\\
& & 720 & \textbf{0.454} & \textbf{0.436} & 0.459 & 0.439 & \underline{0.456} & 0.438 & 0.477 & 0.448 & 0.469 & 0.441 & 0.462 & \underline{0.436}\\\cline{2-15}

& \multirow{4}{*}{\rotatebox{90}{ETTm2}}& 96 & 0.172 & 0.254 & 0.175 & 0.256 & 0.177 & 0.257 & \textbf{0.171} & \textbf{0.253} & 0.176 & 0.257 & \underline{0.172} & \underline{0.253}\\
& & 192 & \underline{0.238} & \underline{0.298} & 0.241 & 0.300 & 0.243 & 0.302 & 0.245 & 0.303 & 0.245 & 0.303 & \textbf{0.238} & \textbf{0.297}\\
& & 336 & 0.299 & \underline{0.337} & \underline{0.299} & 0.337 & 0.302 & 0.342 & 0.304 & 0.341 & 0.306 & 0.342 & \textbf{0.298} & \textbf{0.337}\\
& & 720 & \textbf{0.398} & \underline{0.397} & 0.400 & 0.398 & 0.401 & 0.400 & 0.416 & 0.405 & 0.402 & 0.399 & \underline{0.400} & \textbf{0.396}\\\cline{2-15}

& \multirow{4}{*}{\rotatebox{90}{Exchange}}& 96 & \textbf{0.082} & \textbf{0.201} & 0.091 & 0.210 & 0.090 & 0.209 & 0.092 & 0.212 & \underline{0.087} & \underline{0.206} & 0.088 & 0.208\\
& & 192 & \textbf{0.179} & \textbf{0.301} & 0.190 & 0.309 & 0.184 & 0.306 & 0.188 & 0.308 & 0.185 & 0.306 & \underline{0.183} & \underline{0.305}\\
& & 336 & \textbf{0.333} & \textbf{0.418} & 0.355 & 0.432 & \underline{0.340} & \underline{0.423} & 0.369 & 0.437 & 0.354 & 0.432 & 0.353 & 0.430\\
& & 720 & \textbf{0.846} & \textbf{0.694} & \underline{0.877} & \underline{0.707} & 0.890 & 0.711 & 1.109 & 0.785 & 0.883 & 0.712 & 0.888 & 0.713\\\cline{2-15}

& \multirow{4}{*}{\rotatebox{90}{Weather}}& 96 & \underline{0.159} & \textbf{0.199} & 0.170 & 0.207 & 0.169 & 0.207 & 0.159 & 0.202 & 0.176 & 0.213 & \textbf{0.158} & \underline{0.202}\\
& & 192 & \textbf{0.205} & \textbf{0.245} & 0.217 & 0.251 & 0.217 & 0.251 & 0.209 & 0.248 & 0.222 & 0.255 & \underline{0.207} & \underline{0.246}\\
& & 336 & \textbf{0.265} & \textbf{0.288} & 0.274 & 0.293 & 0.274 & 0.293 & 0.267 & 0.291 & 0.277 & 0.295 & \underline{0.267} & \underline{0.289}\\
& & 720 & \textbf{0.344} & \textbf{0.341} & 0.354 & 0.345 & 0.355 & 0.346 & 0.348 & 0.344 & 0.358 & 0.348 & \underline{0.346} & \underline{0.342}\\\hline

\multicolumn{3}{c}{\# Top-2}
& \multicolumn{2}{c}{45}
& \multicolumn{2}{c}{10}
& \multicolumn{2}{c}{12}
& \multicolumn{2}{c}{2}
& \multicolumn{2}{c}{6}
& \multicolumn{2}{c}{21}
\\\hline

\end{tabular}}
\label{complete_results_multipatchformer}
\end{table}

\bibliographystyle{IEEEtran}
\bibliography{refs}

\end{document}